%% file: main.tex
\documentclass{article}

\usepackage[final]{neurips}

\usepackage{marvosym} %
\usepackage[utf8]{inputenc} %
\usepackage[T1]{fontenc}    %
\usepackage{hyperref}       %
\usepackage{url}            %
\usepackage{booktabs}       %
\usepackage{amsfonts}       %
\usepackage{microtype}      %
\usepackage{xcolor}         %
\usepackage{epigraph}
\usepackage[export]{adjustbox}
\usepackage{multirow}
\usepackage{listings}
\usepackage{colortbl}
\usepackage{color}
\usepackage{pifont}
\usepackage{fontawesome}
\usepackage{wrapfig}
\usepackage{subcaption}
\usepackage{caption}
\usepackage{float}

\usepackage{minitoc} %

\noptcrule

\input{math_commands}

\input{macros}

\title{\model{}: Not All Tokens Are What You Need}

\author{%
  Zhenghao Lin\thanks{Equal contribution. See author contributions for details. Work done during their internships at Microsoft Research Asia. \Letter:~\texttt{zhenghaolin@stu.xmu.edu.cn};~~\texttt{zebgou@gmail.com}}$~~^{\chi\phi}$~~~~
  Zhibin Gou$^{\star\pi\phi}$~~~~
Yeyun Gong\thanks{Correspondence authors.}$~~^{\phi}$~~~~
Xiao Liu$^{\phi}$~~~~
  Yelong Shen$^{\phi}$\\
  \textbf{
Ruochen Xu$^{\phi}$~~~~
Chen Lin$^{\diamond\chi\rho}$~~~~
Yujiu Yang$^{\diamond\pi}$~~~~
Jian Jiao$^{\phi}$~~~~
Nan Duan$^{\phi}$~~~~
Weizhu Chen$^{\phi}$}
\\
$^\chi$Xiamen University\quad
$^\pi$Tsinghua University\quad
$^\rho$Shanghai AI Laboratory\quad \\
$^\phi$Microsoft \\
{\url{https://aka.ms/rho}}
\vspace{-0.5cm}
}

\begin{document}

\doparttoc
\faketableofcontents

\maketitle

\input{sec/0_abs}
\input{sec/1_intro}

\input{sec/2_analysis}

\input{sec/3_method}

\input{sec/4_exps}

\input{sec/7_conclusion}

\input{sec/ack}

{
\small
\bibliographystyle{unsrtnat}
\bibliography{main}
}
\cleardoublepage
\input{sec/8_appendix}

\end{document}

%% file: math_commands.tex
\usepackage{amsmath,amsfonts,bm}

\def\eqref#1{equation~\ref{#1}}

\def\1{\bm{1}}

\DeclareMathAlphabet{\mathsfit}{\encodingdefault}{\sfdefault}{m}{sl}
\SetMathAlphabet{\mathsfit}{bold}{\encodingdefault}{\sfdefault}{bx}{n}

%% file: macros.tex
\usepackage{xspace}
\newcommand{\ie}{\emph{i.e.,}\xspace}
\newcommand{\eg}{\emph{e.g.,}\xspace}
\newcommand{\etc}{\emph{etc}\xspace}

\newcommand{\model}[1]{\textsc{Rho-1}} %
\newcommand{\method}[1]{{SLM}} %
\newcommand{\methodfull}[1]{{Selective Language Modeling}} %

\newcommand{\mamix}[1]{\textsc{MaMix}}
\newcommand{\genmix}[1]{\textsc{GenMix}}

\makeatletter  
\renewcommand\@fnsymbol[1]{%
  \ensuremath{%
    \ifcase#1\or %
    \star\or     %
    \diamond\or   %
    \dagger\or  %
    \mathsection\or %
    \mathparagraph\or %
    \|\or        %
    **\or        %
    \dagger\dagger\or %
    \ddagger\ddagger %
    \else \@ctrerr \fi%
  }%
}  
\makeatother

\definecolor{darkblue}{rgb}{0, 0, 0.5}
\hypersetup{colorlinks=true, citecolor=darkblue, linkcolor=darkblue, urlcolor=darkblue}
\definecolor{lightgray}{rgb}{0.9, 0.9, 0.9}

\definecolor{darkgreen}{RGB}{50,100,0}
\definecolor{darkred}{RGB}{200, 0, 0}
\definecolor{lightred}{RGB}{250, 200, 200}
\definecolor{lightblue}{RGB}{210, 220, 250}
\definecolor{doderblue}{RGB}{30,144,255}
\definecolor{select}{RGB}{222, 235, 247}
\definecolor{unselect}{RGB}{247, 207, 206}

\newcommand{\blue}{\cellcolor{lightblue}}

\newcommand{\grey}{\cellcolor{lightgray}}
\newcommand{\hl}[1]{\textcolor{purple}{#1}}
\newcommand{\hlb}[1]{\textcolor{doderblue}{#1}}

\newcommand{\cmark}{\textcolor{darkgreen}{\ding{51}}} %
\newcommand{\xmark}{\textcolor{darkred}{\ding{55}}} %

\usepackage{cleveref}
\makeatletter
\AtBeginDocument{%
  \renewcommand{\sectionautorefname}{\S\@gobble}

  \renewcommand{\subsectionautorefname}{\S\@gobble}  
}
\makeatother

%% file: sec/0_abs.tex
\begin{abstract}
\label{sec:abstract}
\vspace{-0.2cm}
Previous language model pre-training methods have uniformly applied a next-token prediction loss to all training tokens.
Challenging this norm, we posit that \textit{``Not all tokens in a corpus are equally important for language model training''}.
Our initial analysis examines token-level training dynamics of language model, revealing distinct loss patterns for different tokens. 
Leveraging these insights, we introduce a new language model called \model{}. Unlike traditional LMs that learn to predict every next token in a corpus, \model{} employs Selective Language Modeling (SLM), which selectively trains on useful tokens that aligned with the desired distribution.
This approach involves scoring tokens using a reference model, and then training the language model with a focused loss on tokens with higher scores.
When continual pretraining on 15B OpenWebMath corpus, \model{} yields an absolute improvement in few-shot accuracy of up to 30\% in 9 math tasks.
After fine-tuning, \model{}-1B and 7B achieved state-of-the-art results of 40.6\% and 51.8\% on MATH dataset, respectively — matching DeepSeekMath with only 3\% of the pretraining tokens.
Furthermore, when continual pretraining on 80B general tokens, \model{} achieves 6.8\% average enhancement across 15 diverse tasks, increasing both data efficiency and performance of the language model pre-training.
\end{abstract}

%% file: sec/1_intro.tex
\begin{figure}[h]
\centering
\includegraphics[width=0.95\textwidth]{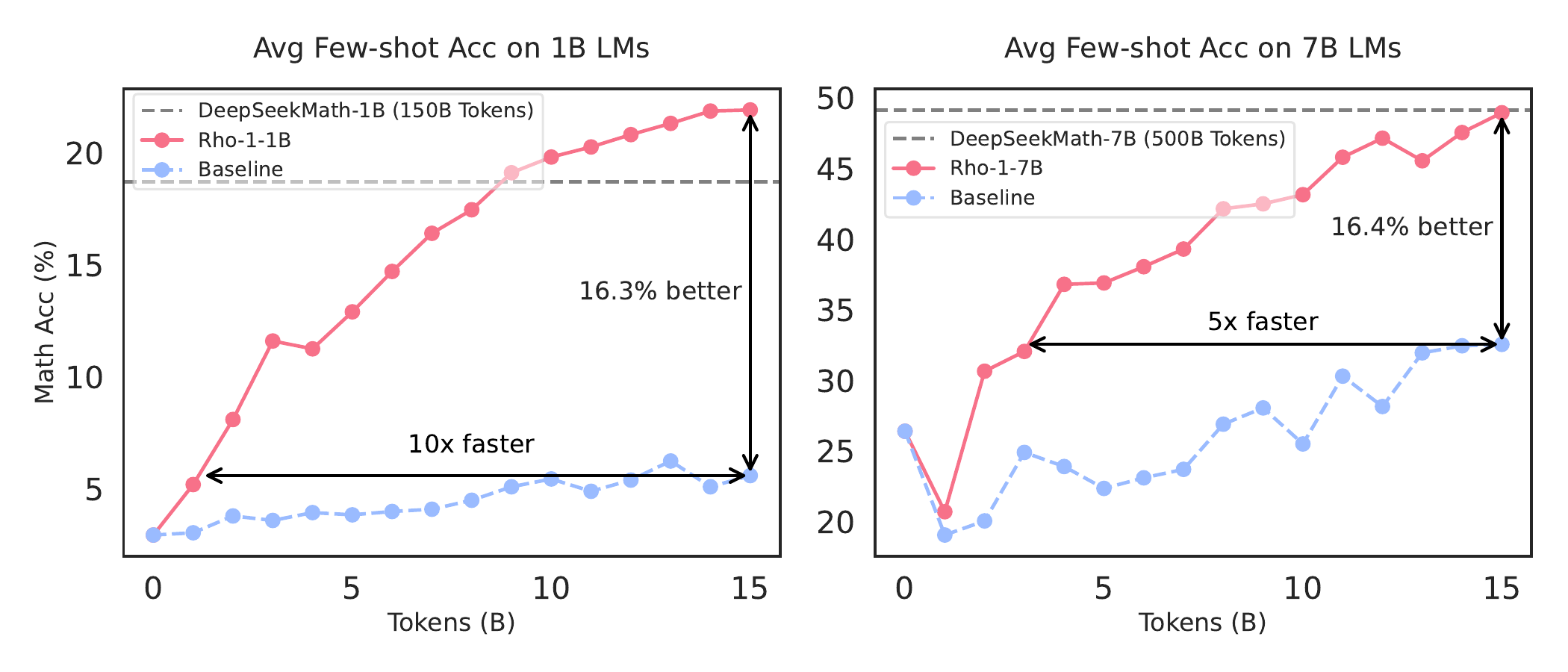}
\caption{
We continual pretrain 1B and 7B LMs with 15B OpenWebMath tokens. \model{} is trained with our proposed \methodfull{} (SLM), while baselines are trained using causal language modeling.
SLM improves average few-shot accuracy on GSM8k and MATH by over 16\%, achieving the baseline performance 5-10x faster.
}
\label{fig:acc_vs_tokens}
\end{figure}

\begin{figure*}[t]
\centering
\includegraphics[width=\textwidth]{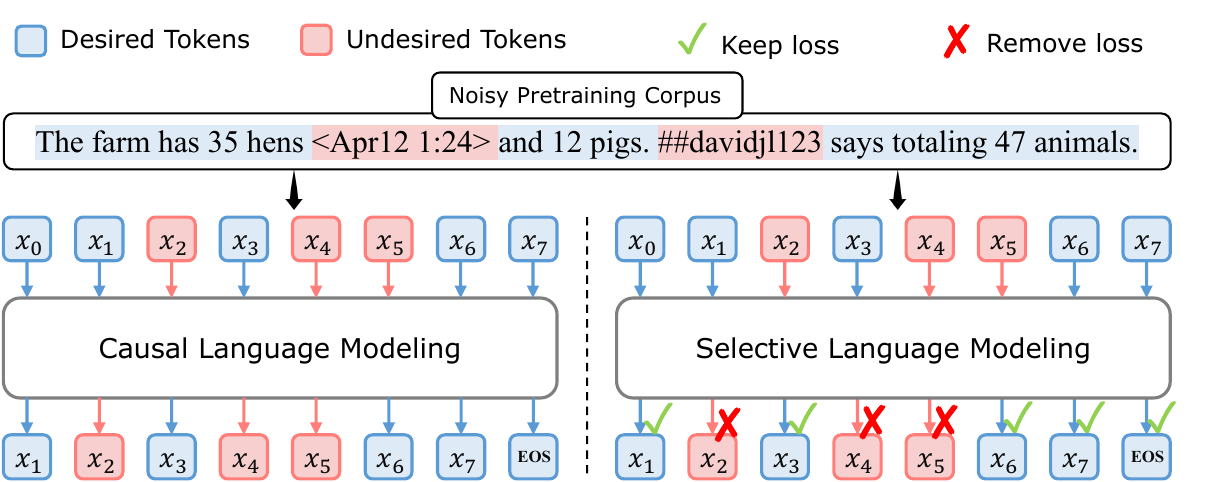}
\caption{
\textbf{Upper:} Even an extensively filtered pretraining corpus contains token-level noise.
\textbf{Left:} Previous Causal Language Modeling (CLM) trains on all tokens.
\textbf{Right:} Our proposed Selective Language Modeling (SLM) selectively applies loss on those useful and clean tokens.
}
\label{fig:token_example}
\end{figure*}

\newpage
\section{Introduction}
\label{sec:introduction}
Scaling up model parameters and dataset size has consistently elevated the next-token prediction accuracy in large language models, yielding significant advancements in artificial intelligence \citep{kaplan2020scaling, brown2020language, openai2023gpt4, team2023gemini}.
However, training on all available data is not always optimal or feasible. As a result, the practice of data filtering has become crucial, using various heuristics and classifiers \citep{brown2020language, wenzek2019ccnet} to select training documents.
These techniques significantly improve data quality and boost model performance.

However, despite thorough document-level filtering, high-quality datasets still contain many noisy tokens that can negatively affect training, as illustrated in \autoref{fig:token_example} (Upper).
Removing such tokens might alter the text's meaning, while overly strict filtering could exclude useful data \citep{welbl2021challenges, muennighoff2024scaling} and lead to biases \citep{dodge2021documenting, longpre2023pretrainer}.
Furthermore, research indicates that the distribution of web data does not inherently align with the ideal distribution for downstream applications~\citep{tay2022scaling,wettig-etal-2023-mask}.
For example, common corpus at the token level may include undesirable content like hallucinations or highly ambiguous tokens that are hard to predict.
Applying the same loss to all tokens can lead to inefficient computation on non-essential tokens, potentially restricting LLMs from achieving more advanced levels of intelligence.

To explore how language models learn at the token level, we initially examined training dynamics, particularly how the token-level loss evolves during usual pretraining.
In \autoref{sec:analysis:dynamics}, we evaluated the model's token perplexity at different checkpoints and categorized tokens into different types.
Our findings reveal that significant loss reduction is limited to a select group of tokens.
Many tokens are ``easy tokens'' that are already learned, and some are ``hard tokens'' that exhibit variable losses and resist convergence. These tokens can lead to numerous ineffective gradient updates.

Based on these analyses, we introduce \model{} models trained with a novel \methodfull{} (SLM) objective \footnote{``Rho'' denotes selective modeling of tokens with higher information ``density ($\rho$)''.}.
As shown in \autoref{fig:token_example} (Right), this approach inputs the full sequence into the model and selectively removes the loss of undesired tokens. 
The detailed pipeline is depicted in~\autoref{fig:pipeline}:
First, SLM trains a reference language model on high-quality corpora. This model establishes utility metrics to score tokens according to the desired distribution, naturally filtering out unclean and irrelevant tokens.
Second, SLM uses the reference model to score each token in a corpus using its loss (\autoref{sec:method:ref}).
Finally, we train a language model only on those tokens that exhibit a high excess loss between the reference and the training model, selectively learning the tokens that best benefit downstream applications (\autoref{sec:method:slm}).

We show through comprehensive experiments that SLM significantly enhances token efficiency during training and improves performance on downstream tasks. Furthermore, our findings indicate that SLM effectively identifies tokens relevant to the target distribution, resulting in improved perplexity scores on benchmarks for models trained with the selected tokens.
\autoref{sec:exp:math_result} shows the effectiveness of SLM on math continual pretraining: both 1B and 7B \model{} outperform CLM-trained baselines by over 16\% on the GSM8k and MATH datasets. SLM reaches baseline accuracy up to 10x faster, as shown in \autoref{fig:acc_vs_tokens}.
Remarkably, \model{}-7B matches the state-of-the-art performance of DeepSeekMath-7B using only 15B tokens, compared to the 500B tokens required by DeepSeekMath.
Upon fine-tuning, \model{}-1B and 7B achieve 40.6\% and 51.8\% on MATH, respectively. 
Notably, \model{}-1B is the first 1B LM to exceed 40\% accuracy, nearing the early GPT-4's CoT performance of 42.5\%.
\autoref{sec:exp:general_result} confirms the efficacy of SLM in general continual pretraining:
Training Tinyllama-1B on 80B tokens with SLM improves 6.8\% on average across 15 benchmarks, with gains over 10\% in code and math tasks.
In \autoref{sec:exp:self_ref}, we demonstrate that in settings without high-quality reference data, we can use SLM for self-referencing, leading to an average improvement of up to 3.3\% in downstream tasks.

%% file: sec/2_analysis.tex
\begin{figure*}[t]
\centering
\includegraphics[width=\textwidth]{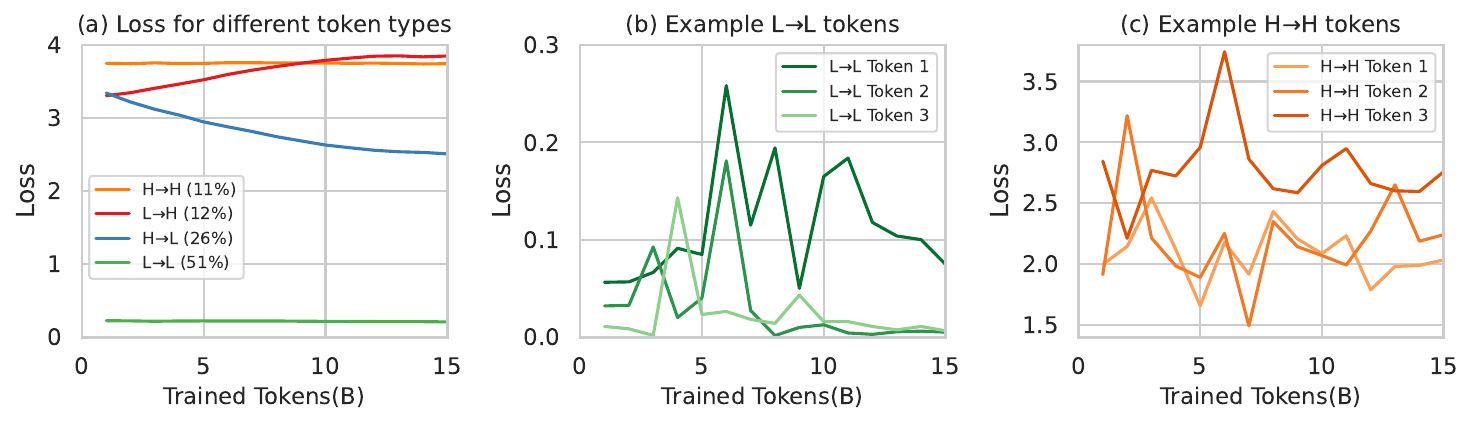}
\caption{\textbf{The loss of four categories of tokens during pretraining.} (a) shows the loss of H→H, L→H, H→L, and L→L tokens during pretraining. (b) and (c) show three cases of fluctuating tokens' loss in L→L and H→H during pretraining, respectively.}
\label{fig:tokenppl}
\end{figure*}

\section{\methodfull{}}

\subsection{Not All Tokens Are Equal: Training Dynamics of Token Loss}
\label{sec:analysis:dynamics}

Our investigation begins with a critical look at how individual tokens' losses evolve during standard pre-training. We continue pre-training Tinyllama-1B with 15B tokens from OpenWebMath, saving checkpoints after every 1B tokens. We then evaluate token-level loss at these intervals using the validation set of approximately 320,000 tokens.
\autoref{fig:tokenppl}(a) reveals a striking pattern: tokens fall into four categories based on their loss trajectory—persistent high loss (H→H), increasing loss (L→H), decreasing loss (H→L), and consistent low loss (L→L). For further details on these categories, see \autoref{sec:appendix:categorie_detail}.
Our analysis uncovers that a mere 26\% of tokens show a notable loss reduction (H→L), while the majority (51\%) remain in the L→L category, indicating they have already been learned.
Interestingly, 11\% of the tokens are persistently challenging (H→H), likely due to high aleatoric uncertainty \citep{hullermeier2021aleatoric}.
Additionally, 12\% of tokens experience an unexpected loss increase (L→H) during training.

Our second observation is that a significant number of token losses exhibit persistent fluctuations, and resist convergence. The loss of many L→L and H→H tokens, as depicted in \autoref{fig:tokenppl} (b) and (c), show high variance during training.
In \autoref{sec:appendix:non_converging_token}, we visualize and analyze the content of these tokens and find that many of them are noisy, which is consistent with our hypothesis.

Consequently, we learn that the loss associated with each token during training does not decrease smoothly like the overall loss; instead, there is a complex training dynamic among different tokens.
If we can select the appropriate tokens for the model to focus on during training, we may be able to stabilize the trajectory of the model's training and enhance its data efficiency.

%% file: sec/3_method.tex
\subsection{Selective Language Modeling}
\label{sec:method:slm}

\begin{figure*}[t]
\centering
\includegraphics[width=\textwidth]{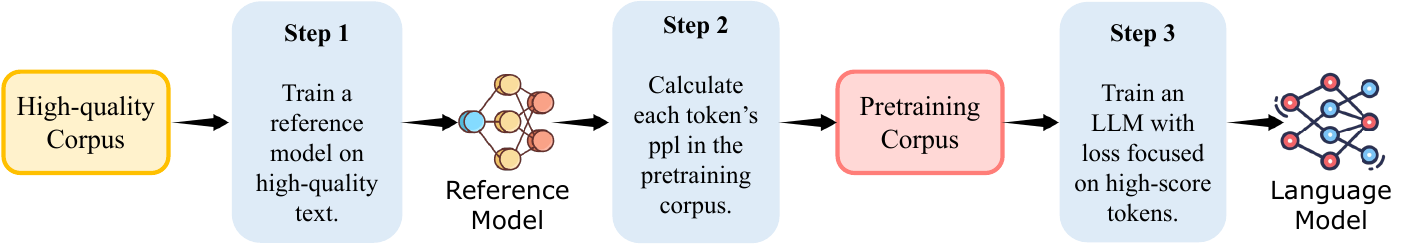}
\caption{\textbf{The pipeline of \methodfull{} (SLM).}
SLM optimizes language model performance by concentrating on valuable, clean tokens during pre-training.
It involves three steps:
(Step 1) Initially, train a reference model on high-quality data.
(Step 2) Then, score each token's loss in a corpus using the reference model.
(Step 3) Finally, selectively train the language model on tokens that have higher scores.
}
\label{fig:pipeline}
\end{figure*}

\paragraph{Overview}
Inspired by the practice of reference model in document-level filtering, we propose a simple pipeline of token-level data selection, termed ``Selective Language Modeling (SLM)''.
Our method comprises three steps, as depicted in \autoref{fig:pipeline}.
We begin by training a reference model on a curated, high-quality dataset. This model then assesses the loss of each token within the pretraining corpus. 
In the final phase, we train the language model selectively, focusing on tokens with high excess loss between the training and reference model.
The intuition is that tokens with high excess loss are more learnable and better aligned with the desired distribution, naturally excluding tokens that are either irrelevant or of low quality.
Below, we provide a detailed description of each step.

\paragraph{Reference Modeling}
\label{sec:method:ref}
We begin by curating a high-quality dataset that reflects the desired data distribution.
We train a reference model (RM) using standard cross-entropy loss on the curated data. The resulting RM is then used to assess the token loss within a larger pretraining corpus.
We compute the reference loss ($\mathcal{L}_{\text{RM}}$) of a token $x_i$ based on the probability that the RM assigns to this token. The calculation is formalized as follows:

\begin{equation}
\mathcal{L}_{\text{RM}}(x_i) = -\log P(x_i | x_{<i})
\label{equ:ref_loss}
\end{equation}

By evaluating $\mathcal{L}_{\text{RM}}$ for each token, we establish the reference loss for selective pretraining, allowing us to focus on the most influential tokens in language modeling.

\paragraph{Selective Pretraining}

Note that Causal Language Modeling (CLM) employs the cross-entropy loss:

\begin{equation}
\mathcal{L}_{\text{CLM}}(\theta) = -\frac{1}{N}\sum_{i=1}^{N} \log P(x_i | x_{<i}; \theta)
\end{equation}

Here, $\mathcal{L_{\text{CLM}}}(\theta)$ represents the loss function parameterized by model $\theta$. $N$ is the length of the sequence, $x_i$ is the $i$-th token in the sequence, and $x_{<i}$ represents all tokens before the $i$-th token.
In contrast, Selective Language Modeling (SLM) trains the language model with a focus on tokens that exhibit a high excess loss when compared to the reference model.
The excess loss ($\mathcal{L}_{\Delta}$) for a token $x_i$ is defined as the difference between the current training model loss ($\mathcal{L}_{\theta}$) and the reference loss:

\begin{equation}
\mathcal{L}_{\Delta}(x_i) = \mathcal{L}_{\theta}(x_i) - \mathcal{L}_{\text{RM}}(x_i)
\end{equation}

We introduce a token selection ratio $k\%$, which determines the proportion of tokens to be included based on their excess loss.
The cross-entropy loss for the selected tokens is computed as follows:

\begin{equation}
\mathcal{L}_{\text{SLM}}(\theta) = -\frac{1}{N*k\%} \sum_{i=1}^{N} I_{k\%}(x_i) \cdot \log P(x_i | x_{<i};\theta)
\end{equation}

Here, $N*k\%$ defines the number of tokens that fall within the top $k\%$ of excess loss. The indicator function $I_{k\%}(x_i)$ is defined as:

\begin{equation}
I_{k\%}(x_i) =
\begin{cases}
1 & \text{if } x_i \text{ ranks in the top } k\% \text{ by } S(x_i) \\
0 & \text{otherwise}
\end{cases}
\end{equation}

By default, we use $\mathcal{L}_{\Delta}$ as the score function $S$.
This ensures that the loss is applied only to the tokens that are deemed most beneficial for the language model to learn from.
In practice, token selection can be implemented by ranking the tokens in a batch according to their excess loss and using only the top $k\%$ of tokens for training.
This process eliminates the loss for undesired tokens without incurring additional costs during pretraining, making our approach both efficient and easily integrated.

%% file: sec/4_exps.tex
\section{Experiments}

We continually pretrained models in both mathematical and general domain and designed ablation and analysis experiments to understand the effectiveness of \method{}.

\subsection{Experimental Setup}
\label{sec:exp:experimental_setup}

\paragraph{Reference Model Training}
To train our mathematical reference model, we gathered a dataset of 0.5B high-quality, math-related tokens. This dataset is a blend of synthetic data from GPT \citep{yu2024metamath,huang2024key} and manually curated data \citep{yue2024mammoth,ni2024exploring}.
For the general reference model, we compiled a corpus of 1.9B tokens from open-source datasets, such as Tulu-v2 \citep{ivison2023camels} and OpenHermes-2.5 \citep{OpenHermes25}.
We trained the reference models for 3 epochs. The maximum learning rate was set at 5e-5 for 1B models and 1e-5 for 7B models, applying a cosine decay schedule.
We set the maximum sequence lengths to 2048 for 1B models and 4096 for 7B models, packing multiple samples into these lengths for model input.
In all main experiments, we initialized the continual pretraining model and the reference model with the \emph{same} base model.

\input{table/main_math_results}
\paragraph{Pretraining Corpus}

For mathematical reasoning, we utilize the OpenWebMath (OWM) dataset \citep{paster2023openwebmath}, which comprises approximately 14B tokens sourced from math-related web pages in the Common Crawl. In the general domain, we combine the SlimPajama \citep{cerebras2023slimpajama} and StarCoderData \citep{starcoder2023} (both part of the Tinyllama corpus) with OpenWebMath, training on a total of 80 billion tokens with a mix ratio of 6:3:1.

\paragraph{Pretraining Setting}
For math pretraining, we continue pretraining on the Tinyllama-1.1B model \citep{zhang2024Tinyllama} and the Mistral-7B model \citep{jiang2023mistral} with learning rates of 8e-5 and 2e-5, respectively.
For the 1.1B model, we conducted our training on 32 × H100 80G GPUs. This configuration allowed us to train approximately 15 billion tokens in around 3.5 hours and 50 billion tokens in about 12 hours. In the case of the 7B model, training the same 15 billion tokens took approximately 18 hours under similar hardware conditions.
For general domain, we set the learning rate for Tinyllama-1.1B model to 1e-4 and train 80B tokens under the same hardware conditions, which takes approximately 19 hours.
The batch size is uniformly set to 1M tokens for both domains. Regarding the token selection ratio, we use 60\% for the Tinyllama-1.1B model and 70\% for the Mistral-7B model.

\paragraph{Baseline Setting}

We use models that have been continually pretrained (Tinyllama-CT and Mistral-CT) through regular causal language modeling as baselines.
Moreover, we compare \model{} with well-known and top-performing baselines, including Gemma~\citep{team2024gemma}, Qwen1.5~\citep{bai2023qwen}, Phi-1.5~\citep{li2023textbooks}, DeepSeekLLM~\citep{deepseek-llm}, DeepSeekMath~\citep{shao2024deepseekmath}, CodeLlama~\citep{roziere2023code}, Mistral~\citep{jiang2023mistral}, Minerva~\citep{lewkowycz2022solving}, Tinyllama~\citep{zhang2024Tinyllama}, LLemma~\citep{azerbayev2023llemma}, and InternLM2-Math~\citep{ying2024internlm}.
For fine-tuning results, we also compare with previous best models MAmmoTH\citep{yue2024mammoth} and ToRA\citep{gou2023tora}.

\paragraph{Evaluation Setup}
To comprehensively evaluate pretrained models, we compare their few-shot capabilities and fine-tuning performance across a variety of tasks.
We adopt the lm-eval-harness\footnote{\url{https://github.com/EleutherAI/lm-evaluation-harness}}~\citep{eval-harness} for general tasks,
and develop math evaluation suite\footnote{\url{https://github.com/ZubinGou/math-evaluation-harness}} for math tasks.
We use vllm (v0.3.2)~\citep{kwon2023efficient} to speed up inference. Further details on our evaluation can be found in \autoref{sec:appendix:evalution}.

\input{table/math_tora_results}

\subsection{Math Pre-training Results}
\label{sec:exp:math_result}

\paragraph{Few-shot CoT Reasoning Results}
We evalute base models prompting with few-shot chain-of-thought (CoT) \citep{wei2022chain} examples following previous works \citep{lewkowycz2022solving, azerbayev2023llemma, shao2024deepseekmath}.
As results shown in \autoref{tab:math-cot-result}, in comparison to continue pretraining directly, \model{}-Math has achieved the average few-shot accuracy improvement of 16.5\% on 1B models and 10.4\% on 7B models. Furthermore, after training for multiple epochs on OpenWebMath, we find that \model{} could further increase the average few-shot accuracy to 40.9\%.
Compared to DeepSeekMath-7B, which pretrained on 500 billion math-related tokens, \model{}-7B pretrained on only 15 billion tokens (selecting 10.5 billion tokens) achieved comparable results, demonstrating the efficiency of our approach.

\paragraph{Tool-Integrated Reasoning Results}
We fine-tune \model{} and baseline models on 69k ToRA corpus \citep{gou2023tora}, consisting of 16k GPT-4-generated trajectories in a tool-integrated reasoning format, and 53k answer-augmented samples using LLaMA.
As presented in \autoref{tab:math-tora}, \model{}-1B and \model{}-7B achieved a state-of-the-art 40.6\% and 51.8\% on MATH dataset, respectively.
On some unseen tasks (\eg TabMWP and GSM-Hard), \model{} also demonstrates a certain degree of generalizability,
with an average few-shot accuracy improvement of 6.2\% on the \model{}-Math-1B and 2.7\% on \model{}-Math-7B.

\begin{figure}[h]
\centering
\includegraphics[width=1.0\textwidth]{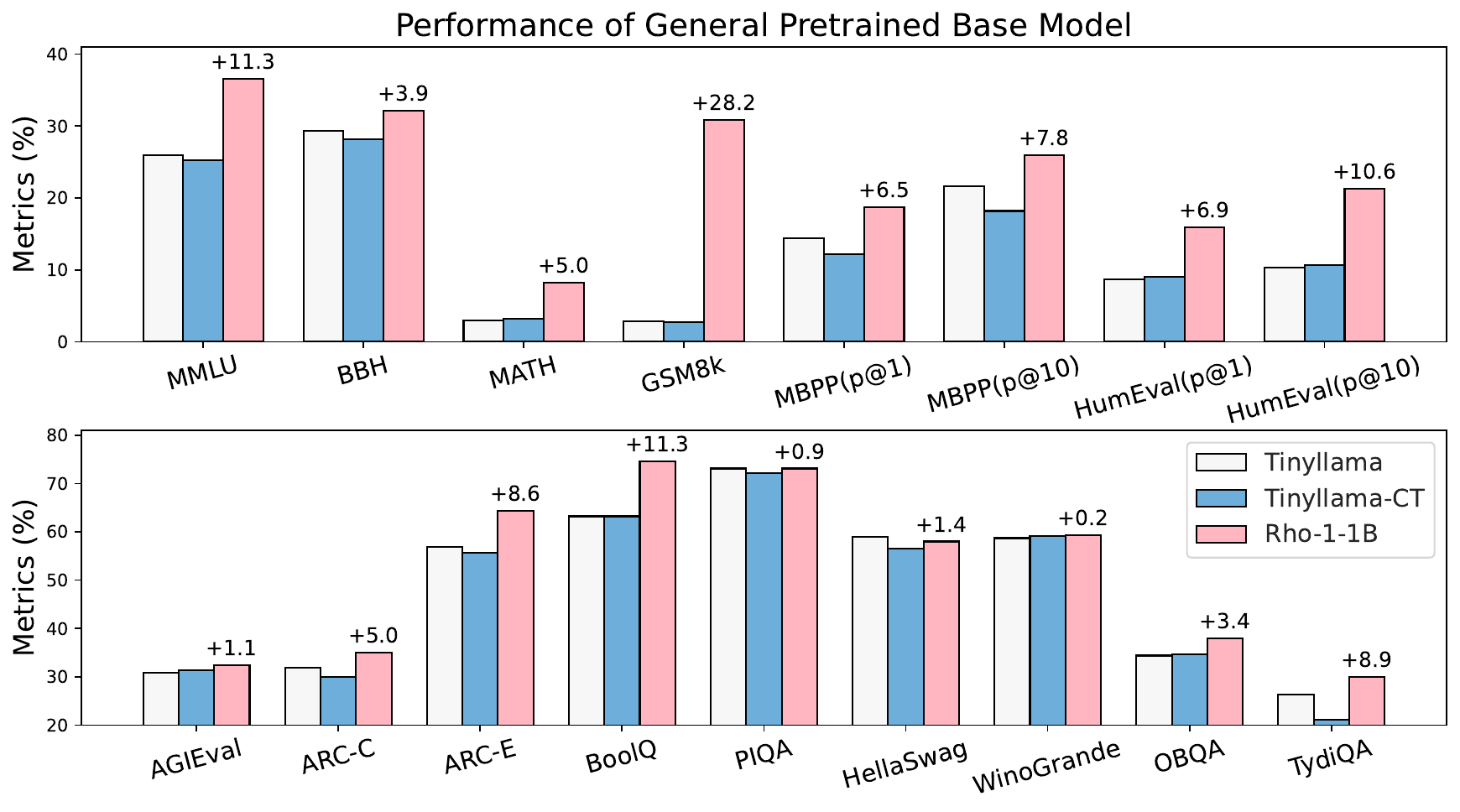}
\caption{
\textbf{General pretraining results.}
We continual pretraining Tinyllama-1B on 80G general tokens. Tinyllama-CT is etrained with CLM, while \model{} is trained with our proposed \method{}.
}
\label{fig:general_result}
\end{figure}

\subsection{General Pre-training Results}
\label{sec:exp:general_result}

We confirm the efficacy of the \method{} in general pretraining by continual training Tinyllama-1.1B on 80 billion tokens.
The results depicted in \autoref{fig:general_result} indicate that although Tinyllama has already undergone extensive training on the majority of these tokens, the application of \method{} yields an average enhancement of 6.8\% across 15 benchmarks compared to direct continual pretraining.
The improvements were especially pronounced in code and math tasks, exceeding 10\%.

\subsection{Self-Reference Results}
\label{sec:exp:self_ref}
\input{table/self_ref_results}

In this section, we demonstrate that SLM can enhance the effectiveness of model pre-training using only pre-training corpora, without the need for additional high-quality data.
Specifically, we initially trained the reference model on the OpenWebMath (OWM) corpus, a subset of Proof-Pile-2 (PPile). We evaluated OWM and PPile using the trained reference model and selected tokens for training. In this scenario, we assume the absence of downstream task-related data, a common situation in real-world applications. We hypothesize that the key factor is not scoring the desired distribution but filtering out noisy tokens.
Therefore, we employed two different scoring functions based on the reference model loss, $\mathcal{L}_\text{RM}$, and the information entropy of the next token, $\mathcal{H}_\text{RM}$, which measures the uncertainty of the next token. Details are provided in \autoref{sec:appendix:self-reference-setting}.

The experimental results, as shown in \autoref{tab:self-ref-1b-ppile}, indicate that using only the OWM-trained reference model can effectively guide the model in pre-training on the same corpus, improving average downstream performance by +2.4\%. Using only the information entropy as the score function brought about a similar improvement.
Additionally, we considered training on the intersection of tokens selected by the two scoring functions and found better performance, with a 40\% reduction in tokens and +3.3\% performance.
Furthermore, training the SLM on the PPile, despite only using the OWM subset to train the reference model, still achieved a 1.8\% improvement with 30\% fewer tokens used.
For more details, please refer to \autoref{sec:appendix:self-reference-setting}.

\begin{figure*}[t]
\centering
\includegraphics[width=\textwidth]{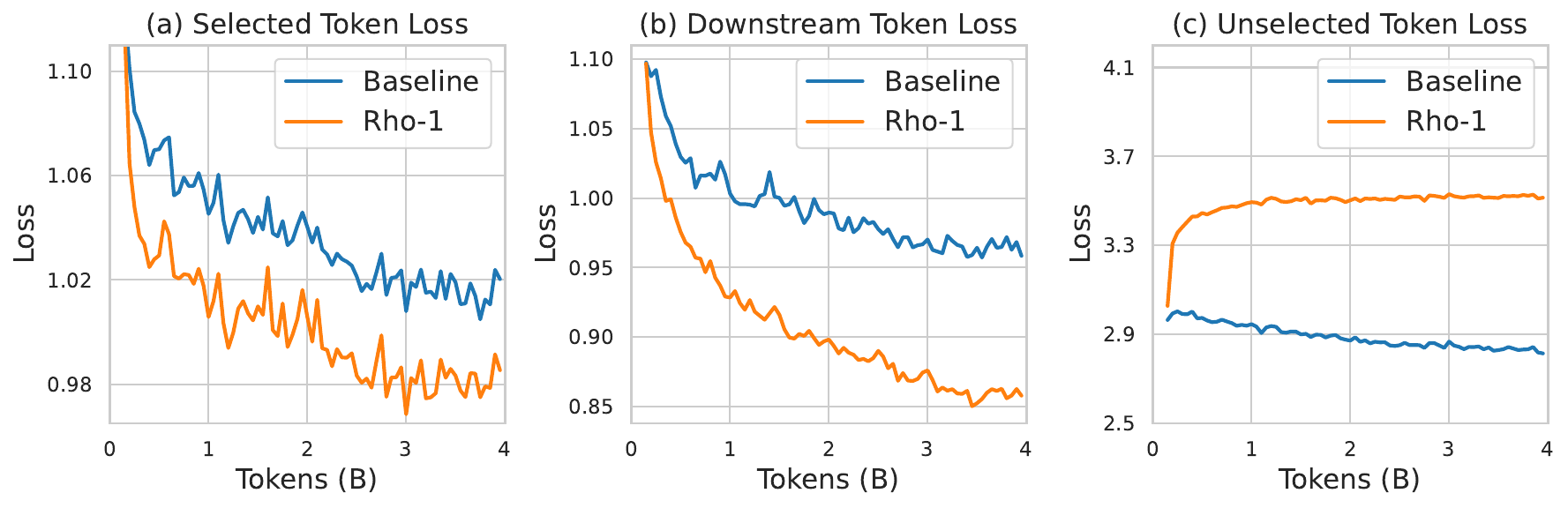}
\caption{\textbf{The dynamics of pretraining loss and downstream loss.} (a) and (c) represent the loss of tokens selected/unselected by \method{} during pretraining in both \method{} and CLM methods, while (b) represents the loss of the \method{} and CLM methods on MetaMath~\citep{yu2024metamath}. We tested the above results through the process of pretraining with a total of 4 billion tokens.}
\label{fig:ppl_select}
\end{figure*}

\subsection{Ablation Study and Analysis}

\paragraph{Selected Token Loss Aligns Better with Downstream Performance}

We utilized the reference model to filter tokens and assess their impact on validation and downstream losses after training. As depicted in \autoref{fig:ppl_select}, we pretrained on 4B tokens and tracked loss variations across methods and validation sets. The \model{} showed greater loss reduction on selected tokens than regular pretraining. Cross-referencing figures (a), (b), and (c) reveals that selected-token pretraining substantially lowers downstream loss, while traditional pretraining's effect on downstream loss is less pronounced despite initial loss reductions. Therefore, we expect that selecting tokens for pretraining is more efficient.

\begin{figure*}[t]
\centering
\includegraphics[width=\textwidth]{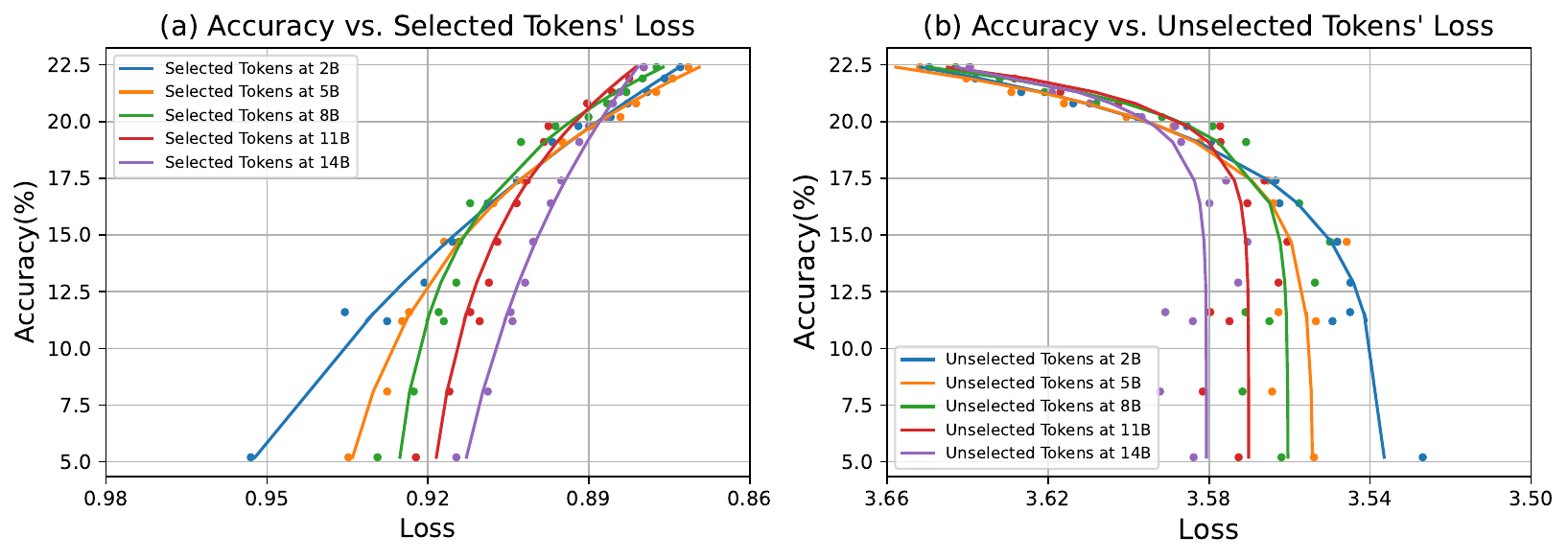}
\caption{\textbf{The relationship between the selected tokens / unselected tokens loss in \method{} and downstream task performance.} The y-axis represents the average few-shot accuracy on GSM8k and MATH. The x-axis represents the average loss on selected tokens / unselected tokens at corresponding checkpoint (2B, 5B, 8B, 11B, and 14B).}
\label{fig:acc_loss}
\end{figure*}

In \autoref{fig:acc_loss}, we demonstrate that the loss of selected tokens correlates with downstream task performance, following a power law similar to recent findings~\citep{gadre2024LanguageMS}. Our analysis shows that tokens selected by \method{} positively impact performance, while those not selected have a negative impact. Thus, reducing loss across all tokens is not imperative for improved model performance. Refer to \autoref{sec:appendix:acc_loss} for further details.

\input{table/ckpt_ppl_and_select_ratio}

\paragraph{What Tokens are Selected with \method{}?}

We aim to analyze the tokens selected by the \method{} method in pretraining to further explore its working mechanism. 
To this end, we visualize the token selection process during the training of \model{} using the OpenWebMath. 
In \autoref{sec:appendix:examples_token_selected}, we have highlighted in \textcolor[RGB]{30,144,255}{blue} the tokens that were retained during actual pretraining. 
We observe that the majority of tokens chosen by the \method{} method are closely related to mathematics, effectively training the model on the parts of the original corpus that are pertinent to mathematical content. 

Furthermore, we investigated the differences in token filtering across various checkpoints during the training process and tested the perplexity of these tokens on different checkpoints. 
As illustrated in \autoref{fig:ckpt_ppl}, we found that the tokens selected by later checkpoints tend to have higher perplexity towards the later stages of training and lower perplexity in the earlier stages.
This may suggest that the model first optimizes tokens with a larger learnable space, thereby increasing learning efficiency.
Moreover, we noticed a sample-wise ``double descent''~\citep{nakkiran2021deep} on the loss of selected tokens, where the select token's perplexity initially increases before decreases.
This might be an effect of selecting tokens based on excess loss, targeting those most in need at each checkpoint.

\paragraph{Effect of Token Select Ratio}
We investigate the impact of token selecting ratios of the \method{}. Generally, the selecting ratio is defined by heuristic rules, similar to the approach previously employed in the training of Masked Language Models (MLMs)~\citep{DevlinCLT19, roberta2019}.
As shown in \autoref{fig:acc_vs_select_ratio}, the selected tokens is suitable for accounting for about 60\% of the original tokens.

%% file: table/main_math_results.tex
\begin{table*}[t]
\centering
\setlength{\tabcolsep}{2pt}
\caption{
\textbf{Few-shot CoT reasoning results of math pretraining.} All models are tested with few-shot prompting. Previous best results are highlighted in \hlb{blue}, while our best results are in \hl{purple}. $^*$Only unique math-related tokens are calculated. For \model{}, we calculate only the selected tokens that are used for training. $^\dagger$We use OpenAI's MATH subset \citep{lightman2023let} for evaluation, since some original test samples have been used in public training sets such as PRM800k. $^\ddagger$The SAT only has 32 four-choice problems, so we average our results over the last three checkpoints, if available.
}
\label{tab:math-cot-result}
\resizebox{\linewidth}{!}{
\begin{tabular}{lrcrr|ccccccccc|c}
\toprule
\textbf{Model} & $|\boldsymbol{\theta}|$ & \textbf{Data} & \multicolumn{1}{c}{\textbf{\begin{tabular}[c]{@{}c@{}}Uniq.\\ Toks$^*$\end{tabular}}}& \multicolumn{1}{c|}{\textbf{\begin{tabular}[c]{@{}c@{}}Train\\ Toks\end{tabular}}} & \textbf{GSM8K} & \textbf{MATH$^\dagger$} & \textbf{SVAMP} & \textbf{ASDiv} & \textbf{MAWPS} & \textbf{TAB}& \textbf{MQA} & \multicolumn{1}{c}{\textbf{\begin{tabular}[c]{@{}c@{}}MMLU\\ STEM\end{tabular}}} & \textbf{SAT$^\ddagger$} & \textbf{AVG} \\
\midrule
\multicolumn{15}{c}{\texttt{1-2B Base Models}} \\
\midrule
\href{https://huggingface.co/Tinyllama/Tinyllama-1.1B-intermediate-step-1431k-3T}{Tinyllama}& 1.1B & - & - & - & 2.9 & 3.2 & 11.0 & 18.1 & 20.4 & 12.5 & 14.6 & 16.1 & 21.9 & 13.4 \\
\href{https://huggingface.co/microsoft/phi-1_5}{Phi-1.5} & 1.3B & - & - & - & 32.4 & 4.2 & 43.4 & 53.1 & 66.2 & 24.4 & 14.3 & 21.8 & 18.8 & 31.0 \\
\href{https://huggingface.co/Qwen/Qwen1.5-1.8B}{Qwen1.5} & 1.8B & - & - & - & \hlb{36.1} & 6.8 & \hlb{48.5} & \hlb{63.6} & \hlb{79.0} & 29.2 & 25.1 & 31.3 & 40.6 & \hlb{40.0} \\
\href{https://huggingface.co/google/gemma-2b}{Gemma} & 2.0B & - & - & - & 18.8 & 11.4 & 38.0 & 56.6 & 72.5 & \hlb{36.9} & \hlb{26.8} & \hlb{34.4} & 50.0 & 38.4 \\ 
DeepSeekLLM & 1.3B & OWM & 14B & 150B & 11.5 & 8.9 & - & - & - & -  & - & 29.6 & 31.3 & -\\
DeepSeekMath & 1.3B & - & 120B & 150B & 23.8 & \hlb{13.6} & - & - & - & - & - & {33.1} & \hlb{56.3} & - \\
\midrule
\multicolumn{15}{c}{\texttt{Continual Pretraining on Tinyllama-1B}}
\\ \midrule
Tinyllama-CT & 1.1B & OWM & 14B & 15B & 6.4 & 2.4 & 21.7 & 36.7 & 47.7 & 17.9 & 13.9 & 23.0 & 25.0 & 21.6 \\
\model{}-Math & 1.1B & OWM & 14B & {9B} & 29.8 & 14.0 & 49.2 & 61.4 & 79.8 & 25.8 & 30.4 & \hl{24.7} & 28.1 & 38.1 \\
$\Delta$  & & & & \grey{-40\%} & \blue{+23.4} & \blue{+11.6} & \blue{+27.5} & \blue{+24.7} & \blue{+32.1} & \blue{+7.9} & \blue{+16.5} & \blue{+1.7} & \blue{+3.1} & \blue{\textbf{+16.5}} \\
\midrule
\model{}-Math & 1.1B & OWM & 14B & {30B} & \hl{36.2} & \hl{15.6} & \hl{52.1} & \hl{67.0} & \hl{83.9} & \hl{29.0} & \hl{32.5} & {23.3} & \hl{28.1} & \hl{40.9} \\
\midrule
\multicolumn{15}{c}{\texttt{$\ge$ 7B Base Models}}
\\ \midrule
\href{https://huggingface.co/meta-llama/Llama-2-7b-hf}{LLaMA-2} & 7B & & - & - & 14.0 & 3.6 & 39.5 & 51.7 & 63.5 & 30.9 & 12.4 & 32.7 & 34.4 & 31.4 \\
\href{https://huggingface.co/mistralai/Mistral-7B-v0.1}{Mistral} & 7B & & - & - & 41.2 & 11.6 & 64.7 & 68.5 & 87.5 & 52.9 & 33.0 & 49.5 & 59.4 & 52.0 \\
Minerva & 8B & - & 39B & 164B & 16.2 & 14.1 & -& -& -& -& -& 35.6 & - & - \\
Minerva & 62B & - & 39B & 109B & 52.4 & 27.6 & -& -& -& -& -& 53.9 & - & - \\
Minerva & 540B & - & 39B & 26B & 58.8 & {33.6} & -& -& -& -& -& \hlb{63.9} & - & - \\
\href{https://huggingface.co/EleutherAI/llemma_7b}{LLemma} & 7B & PPile & 55B & 200B & 38.8 & 17.2 & 56.1 & 69.1 & 82.4 & 48.7 & 41.0 & 45.4 & 59.4 & 50.9 \\
\href{https://huggingface.co/EleutherAI/llemma_34b}{LLemma} & 34B & PPile & 55B & 50B & 54.2 & 23.0 & 67.9 & {75.7} & 90.1 & 57.0 & 49.8 & 54.7 & 68.8 & 60.1 \\
\href{https://huggingface.co/internlm/internlm2-math-base-7b}{Intern-Math} & 7B & - & 31B & 125B & 41.8 & 14.4 & 61.6 & 66.8 & 83.7 & 50.0 & 57.3  & 24.8 & 37.5 & 48.7 \\
\href{https://huggingface.co/internlm/internlm2-math-base-20b}{Intern-Math} & 20B & - & 31B & 125B & \hlb{65.4} & 30.0 & \hlb{75.7} & 79.3 & \hlb{94.0} & 50.9 & 38.5 & 53.1 & 71.9 & 62.1\\
\href{https://huggingface.co/deepseek-ai/deepseek-math-7b-base}{DeepSeekMath} & 7B & - & 120B & 500B & {64.1} & \hlb{34.2} & {74.0} & \hlb{83.9} & {92.4} & \hlb{63.4} & \hlb{62.4} & 56.4 & \hlb{84.4} & \hlb{68.4} \\
\midrule
\multicolumn{15}{c}{\texttt{Continual Pretraining on Mistral-7B}} \\
\midrule
Mistral-CT & 7B & OWM & 14B & 15B & 42.9 & 22.2 & 68.6 & 71.0 & 86.1 & 45.1 & 47.7 & 52.6 & 65.6 & 55.8\\
\model{}-Math & 7B & OWM & 14B & {10.5B} & \hl{66.9} & \hl{31.0} & \hl{77.8} & \hl{79.0} & \hl{93.9} & \hl{49.9} & \hl{58.7} & \hl{54.6} & \hl{84.4} & \hl{66.2}\\
$\Delta$  & & & & \grey{-30\%} & \blue{+24.0} & \blue{+8.8} & \blue{+9.2} & \blue{+8.0} & \blue{+7.8} & \blue{+4.8} & \blue{+11.0} & \blue{+2.0} & \blue{+18.8} &  \blue{\textbf{+10.4}}\\  
\bottomrule
\end{tabular}
}
\end{table*}

%% file: table/math_tora_results.tex
\begin{table}[t]
\caption{\textbf{Tool-integrated reasoning results of math pretraining.}}
\label{tab:math-tora}
\centering
\setlength{\tabcolsep}{5pt}
\resizebox{\linewidth}{!}{
\begin{tabular}{lrcc|ccccccc|c}
\toprule
\textbf{Model} & \textbf{Size} & \textbf{Tools} & \textbf{SFT Data} & \textbf{GSM8k} & \textbf{MATH} 
 & \textbf{SVAMP} &  \textbf{ASDiv} & \multicolumn{1}{c}{\textbf{MAWPS}} & \textbf{TAB} & \textbf{GSM-H} & \multirow{2}{*}{\textbf{AVG}} \\
\cmidrule{1-11}
\multicolumn{4}{l|}{\textbf{Used for SFT?}} & \cmark & \cmark & \xmark & \xmark & \xmark & \xmark & \xmark & \\
\midrule
\multicolumn{12}{c}{\texttt{Previous Models}} \\
\midrule
GPT4-0314 & - & \xmark & - &  92.0 & 42.5 & 93.1  & 91.3 & 97.6  & 67.1 & 64.7 & 78.3\\
GPT4-0314 (PAL) & - & \cmark & - & 94.2 & 51.8 & 94.8  & 92.6 & 97.7  & 95.9 & 77.6 & 86.4\\
MAmmoTH & 70B & \cmark & MI-260k & 76.9 & 41.8 & 82.4 & - & - & - & - & - \\
ToRA & 7B & \cmark & ToRA-69k & {68.8} & {40.1} & {68.2}  & {73.9} & {88.8} & 42.4  & {54.6}& {62.4} \\
ToRA & 70B & \cmark & ToRA-69k & {{84.3}} & {{49.7}}  & {{82.7}}  & {{86.8}} & {{93.8}}  & {{74.0}} & {{67.2}}& {{76.9}} \\
DeepSeekMath & 7B & \cmark & ToRA-69k & 79.8 & 52.0 & 80.1 & 87.1 & 93.8 & 85.8 & 63.1 & 77.4  \\
\midrule
\multicolumn{12}{c}{\texttt{Our Pretrained Models}} \\
\midrule
TinyLlama-CT & 1B  &\cmark &  ToRA-69k & 51.4 & 38.4 & 53.4 & 66.7 & 81.7 & 20.5 & 42.8 & 50.7\\
\model{}-Math & 1B & \cmark & ToRA-69k & \hl{59.4} & \hl{40.6} & \hl{60.7} & \hl{74.2} & \hl{88.6} & \hl{26.7} & \hl{48.1} & \hl{56.9}
\\
$\Delta$ &  &  &  & \blue{+8.0} & \blue{+2.2} & \blue{+7.3} & \blue{+7.5} & \blue{+6.9} & \blue{+6.2} & \blue{+5.3} & \blue{\textbf{+6.2}} \\ 
\midrule
Mistral-CT & 7B & \cmark & ToRA-69k & 77.5 & 48.4 & 76.9 & 83.8 & 93.4 & 67.5 & 60.4 & 72.6\\
\model{}-Math & 7B & \cmark & ToRA-69k & \hl{81.3} & \hl{51.8} & \hl{80.8} & \hl{85.5} & \hl{94.5} & \hl{70.1} & \hl{63.1} & \hl{75.3} \\
$\Delta$ &  &  &  & \blue{+3.8} & \blue{+3.4} & \blue{+3.9} & \blue{+1.7} & \blue{+1.1} & \blue{+2.6} & \blue{+2.7} & \blue{\textbf{+2.7}} \\
\bottomrule
\end{tabular}
}
\end{table}

%% file: table/self_ref_results.tex
\begin{table*}[t]
\centering
\small
\setlength{\tabcolsep}{2pt}
\caption{\textbf{Self-Reference results.} We use OpenWebMath (OWM) to train the reference model.}
\label{tab:self-ref-1b-ppile}
\resizebox{\linewidth}{!}{
\begin{tabular}{l|cccc|cccccc|c}
\toprule
\textbf{Model} & \multicolumn{1}{c}{\textbf{\begin{tabular}[c]{@{}c@{}}Score \\ Function \end{tabular}}} & \textbf{Data} & \multicolumn{1}{c}{\textbf{\begin{tabular}[c]{@{}c@{}}Uniq.\\ Toks\end{tabular}}}& \multicolumn{1}{c|}{\textbf{\begin{tabular}[c]{@{}c@{}}Train\\ Toks\end{tabular}}} & \textbf{GSM8K} & \textbf{MATH} & \textbf{SVAMP} & \textbf{ASDiv} & \textbf{MAWPS} & \textbf{MQA} & \textbf{AVG} \\
\midrule
Tinyllama-CT (RM) & - & OWM & 14B & 15B & 6.3 & 2.6 & 21.7 & 36.7 & 47.7 & 13.9 & 21.5 \\
Tinyllama-SLM & $\mathcal{L}_{\text{RM}}$ & OWM & 14B & 10.5B & 6.7 & 4.6 & 23.3 & 40.0 & 54.5 & 14.3 & 23.9  \\
Tinyllama-SLM & $\mathcal{H}_{\text{RM}}$ & OWM & 14B & 10.5B & 7.0 & 4.8 & 23.0 & 39.3 & 50.5 & 13.5 & 23.0 \\
Tinyllama-SLM & $\mathcal{L}_{\text{RM}}\cap\mathcal{H}_{\text{RM}}$ & OWM & 14B & 9B & \hl{7.1} & \hl{5.0} & \hl{23.5} & \hl{41.2} & \hl{53.8} & \hl{18.0} & \hl{24.8} \\
\midrule
Tinyllama-CT & - & PPile & 55B & 52B & 8.0 & 6.6 & 23.8 & 41.0 & 54.7 & 14.2 & 24.7   \\
Tinyllama-SLM & $\mathcal{L}_{\text{RM}}\cap\mathcal{H}_{\text{RM}}$ & PPile & 55B & 36B & 8.6 & 8.4 & 24.4 & 43.6 & 57.9 & 16.1 & 26.5 \\
\bottomrule
\end{tabular}
}
\end{table*}

%% file: table/ckpt_ppl_and_select_ratio.tex
\begin{table}[t]
\begin{minipage}{0.49\linewidth}
  \centering
  \includegraphics[width=1.0\textwidth]{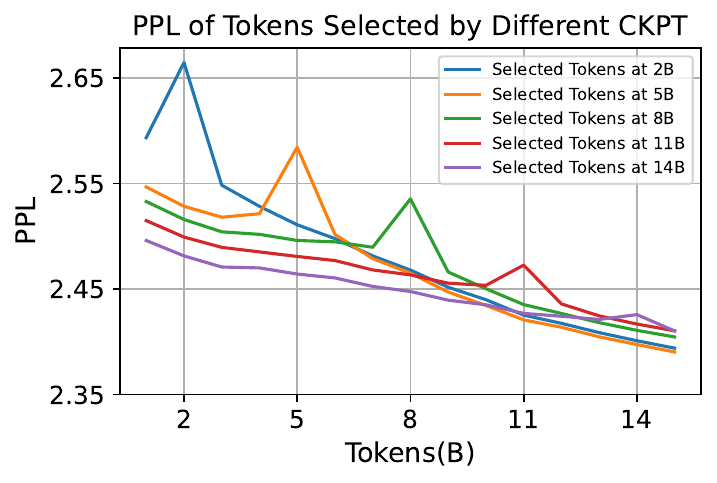}
  \captionsetup{type=figure}
  \captionof{figure}{\textbf{The PPL of tokens selected by different checkpoint.} We test the PPL of the tokens selected at 2B, 5B, 8B, 11B, and 14B.
  }
  \label{fig:ckpt_ppl}
\end{minipage}
\hfill
\begin{minipage}{0.48\linewidth}
  \centering
  \includegraphics[width=1.0\textwidth]{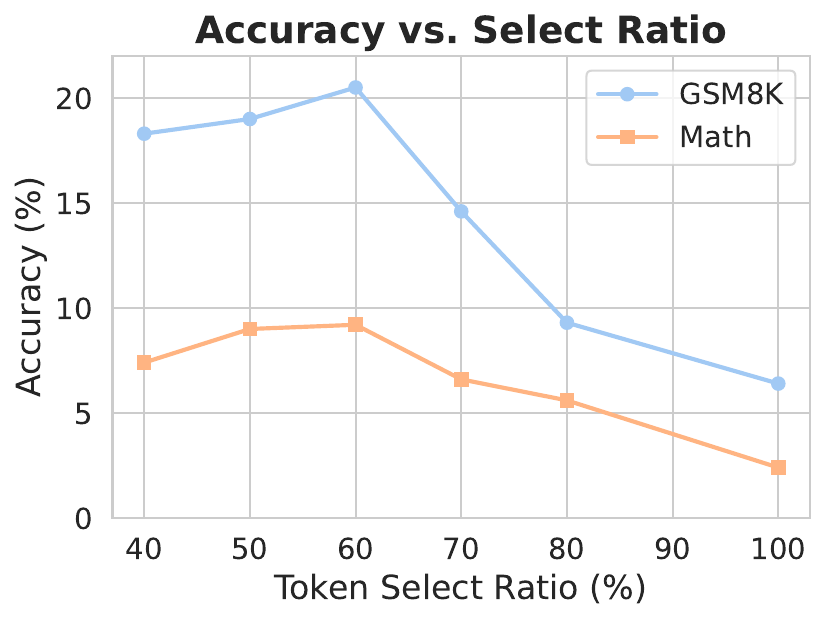}
  \captionsetup{type=figure}
  \captionof{figure}{\textbf{Effect of token select ratio.} We train 1B LM with SLM objective on 5B tokens.}
  \label{fig:acc_vs_select_ratio}
\end{minipage}
\hfill

\end{table}

%% file: sec/7_conclusion.tex
\section{Conclusion}

In this paper, we propose using Selective Language Modeling(SLM) to train \model{}, which select more suitable tokens for current pretraining stage. We conducted the detailed analysis of the loss of tokens during the pretraining process and found that not all tokens are equal during pretraining. Our experiments and analysis in the fields of mathematics and general have demonstrated the effectiveness of the SLM method, emphasizing the importance of token level in the LLM pretraining process. In the future, how to improve pretraining of LLMs from the perspective of token level worthy of in-depth research.

%% file: sec/ack.tex
\subsubsection*{Acknowledgments}

Zhenghao Lin and Chen Lin were supported by National Key R\&D Program of China (No. 2022ZD0160501), the Natural Science Foundation of China (No.62372390,62432011).
Zhibin Gou and Yujiu Yang were supported by the Shenzhen Science and Technology Program (JCYJ20220818101001004) and the ``Graph Neural Network Project'' of Ping An Technology (Shenzhen) Co., Ltd.

%% file: sec/8_appendix.tex
\newpage
\appendix

\part{}
\section*{\centering \LARGE{Appendix}}
\mtcsettitle{parttoc}{Contents}
\parttoc

\clearpage

\input{sec/contrib}

\input{sec/5_related}

\input{sec/6_discussion}

\input{sec/appendix/a1_token_analysis}
\input{sec/appendix/a2_evalution}

\input{sec/appendix/a3_acc_loss}
\input{sec/appendix/a4_examples_tokens_selected}

\input{sec/appendix/a5_self-reference_setting}

\input{sec/appendix/a6_weak-to-strong}

\input{sec/appendix/a7_example_case}

%% file: sec/contrib.tex
\section{Author Contributions}

Zhenghao Lin designed and implemented detailed token selection process, conducted extensive preliminary experiments, developed the pre-training and evaluation pipeline, conducted most of the pre-training experiments and analysis, implemented baselines, and significantly contributed to the writing.
Zhibin Gou presented a preliminary proposal, introduced the method of using excess loss for reweighting tokens, compiled high-quality corpora, trained reference models, set up the fine-tuning and evaluation pipelines, designed the experimental analysis, and significantly contributed to the writing.
Yeyun Gong proposed the initial project and co-led the project with Weizhu Chen, they offered extensive advice and guidance on experiments and writing, and oversaw team collaboration and resource management.
Xiao Liu, Yelong Shen, Ruochen Xu, Chen Lin, Yujiu Yang, Jian Jiao, and Nan Duan offered research mentorship, coordinated the project, and contributed to the writing.

%% file: sec/5_related.tex
\section{Related Works}

\subsection{Pretraining Data Optimization}

The objective of optimizing pre-training corpora is to maximize the performance and efficiency of language model training by improving the quality and scale of the pretrain data mixture.
This includes data collecting through crawling \citep{raffel2020exploring} or synthesis \citep{polu2020generative, gunasekar2023textbooks}, de-duplication \citep{lee2021deduplicating, kandpal2022deduplicating, tirumala2023d4}, filtering and selection \citep{xie2024data, albalak2024survey}, as well as data composition \citep{xie2024doremi} and curriculum \citep{chen2024skill,ma2024at}.

\subsection{Data Selection}
Data selection for fine-tuning has been extensively studied, focusing on improving quality \citep{li2023quantity}, diversity \citep{liu2024makes}, and distribution matching \citep{li2023one, xia2024less, ni2024exploring, kang2024get}.
For pretraining, various lightweight filters are utilized \citep{albalak2024survey}, including heuristic-based (\eg language and item count filtering), classifier-based \citep{brown2020language}, and perplexity or loss-based approaches \citep{wenzek2019ccnet, qin2024infobatch}.
The massive public RedPajama-Data-v2 dataset \citep{together2023redpajama}, for example, leverages over 40 quality indicators for data filtering and reweighting. Nevertheless, strict filtering like blocklist \citep{raffel2020exploring} and Safety API filtering \citep{welbl2021challenges}, have been found to hurt evaluation loss or induce bias \citep{dodge2021documenting}.

Sample-level selection has been extensively studied in previous research \citep{sener2017active,killamsetty2021retrieve}, particularly through online batch selection \citep{loshchilov2015online, schaul2015prioritized, chang2017active, katharopoulos2018not, jiang2019accelerating}. These approaches have been applied to various classification tasks \citep{song2020carpe,2022PrioritizedTraining} and language modeling \citep{fan2023irreducible}.
However, \citet{kaddour2023no} find that batch selection is not computationally efficient.

Many previous works have employed the general idea of using a reference model as a proxy for data selection. For instance, \textit{Selection Via Proxy} trains a proxy model to select samples with high uncertainty \citep{coleman2019selection}.  \citet{xie2024data} and \citet{engstrom2024dsdm} utilize n-gram models or datamodels with a target dataset to estimate importance weights. Additionally, \citet{xie2024doremi} optimize the worst-case excess loss \citep{oren2019distributionally} relative to a reference model to determine domain weights.
One of SLM’s scoring functions is excess loss, and the most relevant work related to excess loss is RHO-LOSS \citep{2022PrioritizedTraining}, which trains a small model on a holdout set and uses the difference between training loss and holdout loss to select in-batch samples.
Although excess loss is mathematically identical to RHO-LOSS, SLM differs in three important ways:
1) The focus is distinct. Motivated by the training dynamics of token loss, the core idea of SLM is to select useful tokens for pre-training. Its score functions are highly flexible and not limited to excess loss (see Appendix \ref{sec:appendix:self-reference-setting} for other functions). In contrast, RHO-LOSS aims to mathematically derive a reducible holdout loss to minimize generalization loss.
2) The meaning and training procedure of the proxy model are different. SLM trains a reference model on high-quality data to reflect the desired data distribution, whereas RHO-LOSS trains a small model on a random holdout set.
3) The selection scale and granularity vary. RHO-LOSS selects sample-level data on a small scale (typically 1K–1M samples) for task-specific fine-tuning tasks such as MNIST \citep{lecun1998gradient} and SST-2 \citep{socher2013recursive}. In contrast, SLM conducts fine-grained token-level selection on large-scale language model pre-training, involving up to 80B tokens.

Token-level training strategies have also been explored, especially for the pre-training of BERT-like models using Masked Language Modeling (MLM) \citep{devlin2018bert}.
Specifically, ``selective masking'' involves masking important tokens in the input to focus on learning tokens that are more relevant to downstream tasks \citep{gu-etal-2020-train, lad2022using}, whereas ``token dropping'' aims to reduce training costs by omitting less important tokens \citep{zhong-etal-2023-revisiting, hou-etal-2022-token}. \citep{li2023error} assesses the quality of each token based on the skewness of its predicted distribution and truncates the noisy tokens during training. 
Additionally, some research has approached the analysis and detection of under-trained tokens from a tokenization perspective \citep{rumbelow2023solidgoldmagikarp, land2024fishing}.
To our knowledge, we are the first to explore token-level data selection for large language model training, aimed at enhancing data quality and information density at the most fundamental granularity.

\subsection{Language Model Training Dynamics}
Investigating the training dynamics of language models is essential for understanding their behavior throughout the training process. This research includes studying internal representations \citep{saphra2018understanding}, the acquisition of linguistic knowledge \citep{choshen2021grammar, liu2021probing}, and the phenomenon of grokking \citep{power2022grokking}. The analysis by \citet{xia2022training} is the most related to ours, which examines token-level training trajectories in models of varying sizes.
Our findings, however, diverge from those of \citet{xia2022training}, who posit that tokens with little change in perplexity are ``already learned''. We identify a spectrum of token patterns, including ``easy tokens'' and ``hard tokens'' that resist convergence. Recognizing this, we propose a method of selective language modeling that targets the influential tokens, optimizing the learning process.

\subsection{Scaling Laws}
Scaling laws guide us in discovering the impact of factors such as parameter count, data size, and compute on language model performance and behavior.
These studies usually focus on predicable scaling though power law \citep{kaplan2020scaling, hernandez2021scaling}, optimal resource allocation \citep{hoffmann2022training}, downstream tasks \citep{wei2022emergent, isik2024scaling, gadre2024LanguageMS}, architectures \citep{tay2022scaling}, memorization \citep{tirumala2022memorization, carlini2022quantifying, henighan2023superposition, biderman2024emergent}, and repeating data \citep{hernandez2022scaling, muennighoff2024scaling, xue2024repeat}.
Most scaling laws on model performance study cross-entory loss on all training tokens, while we focus on the tokens loss of desired distributions.

%% file: sec/6_discussion.tex
\section{Limitations and Future Work}
\label{sec:appendix:discussion_and_future_work}

\paragraph{Generalizability}
In math continual pretraining, as depicted in \autoref{fig:ppl_select}, training exclusively with \method{} leads to quickly convergence to the domain focused by the reference model, accompanied by a significant rise in the loss of unselected tokens.
Although no adverse effects, like biases, have been observed from the increased loss yet, a general pretraining loss on text and code may prevent overfitting \citep{goodhart1984problems}, as suggested by \citet{ouyang2022training} and \citet{azerbayev2023llemma}.
Furthermore, future efforts could broaden the corpus scope of the reference model, and enlarge the pretraining data size, as exemplified by DeepSpeedMath \citep{shao2024deepseekmath}.

\paragraph{Scalability}
Due to budget constraints, we have only verified the effectiveness of our method on smaller models (<=7B parameters) and smaller datasets (<100B tokens). Smaller models benefit significantly from removing the loss of irrelevant tokens and focusing on important ones.
However, it's possible that very large models trained on extensive corpora may naturally develop this inductive bias to compress useful data (\ie{} compressing everything), although it may sounds inefficient for now.
Therefore, future works should study whether this selective language modeling technique can scale to very large models and data \citep{kaplan2020scaling}.

\paragraph{Is training a reference model necessary?}

To score tokens, we need a high-quality reference model. This could be a base model trained with a small amount of high-quality data, or a performant open-source model.
In fact, since we only need input logprobs or perplexity from reference model, we could even utilize more powerful proprietary model APIs. We can input tokens and use the log probabilities of the input returned by the API as reference scores. We leave this for future works.

\paragraph{How to improve upon \method{}?} 
There are many natural extensions of \method{}, \eg
reweighting tokens instead of selecting may improve robustness; using a reference model as a reward model to guide pretraining with reinforcement learning; adopting multiple reference models to reduce overfitting; designing token-level curriculum learning and iterative strategies for continuous improvements, \etc.

\paragraph{Expanding the use of \method{}}
\method{} may be extended to supervised fine-tuning to address the noise and distribution mismatches in many SFT datasets.
Another potential application is alignment, \eg by training a reference model to emphasize helpfulness, truthfulness, and harmlessness, we may obtain a base model that is natively aligned during the pretraining stage.
Meanwhile, we believe that the idea of SLM may find broader applications in multimodal data such as images, videos, and speech, which have a high noise-to-information ratio than text.

%% file: sec/appendix/a1_token_analysis.tex
\section{Analysis and Visualization of Tokens in Pretraining}

\begin{figure*}[t]
\centering
\includegraphics[width=\textwidth]{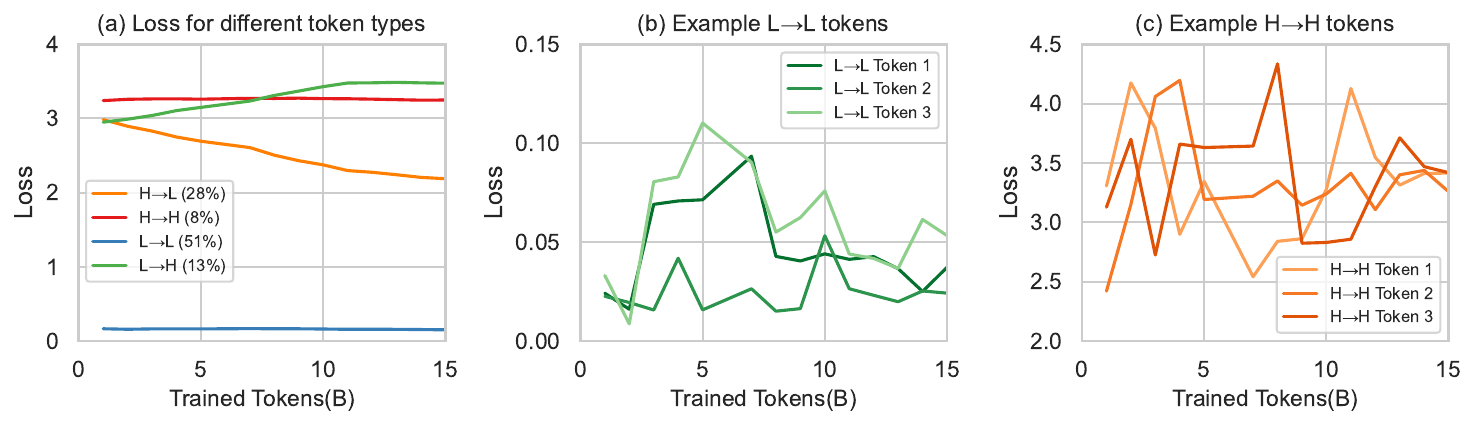}
\caption{\textbf{The loss of four categories of tokens during Mistral-7B pretraining on OpenWebMath.} (a) shows the loss of H→H, L→H, H→L, and L→L tokens during pretraining. (b) and (c) show three cases of fluctuating tokens' loss in L→L and H→H during pretraining, respectively.}
\label{fig:tokenppl_7b}
\end{figure*}

\subsection{More Details of Four Categories Tokens}
\label{sec:appendix:categorie_detail}

We categorize tokens into four categories: H→H, L→H, H→L, L→L. During the training process, we collected the loss of each token after training on each 1 billion tokens training data. We then used linear fitting and took the difference in loss between the first and last points as evidence of whether the loss decreased during the training process.

Specifically, suppose we have a sequence of token's loss $(l_0, l_1, ..., l_n)$. Our goal is to minimize the sum of the squares of the differences between each data point and its linear predictive value:

\begin{equation}
f(a, b) = \text{minimize} \sum_{i=0}^n (l_i - (ax_i + b))^2,
\label{equ:lossfitting}
\end{equation}

where $x_0=0$ is the initial checkpoint and $x_n=n$ is the final checkpoint. Substituting these into the fitted equation, we can obtain the Loss values at the start and end after fitting: $ \mathcal{L}_{\text{start}} = b$ and $\mathcal{L}_{\text{end}} = an + b$. The change in loss can then be expressed as: $\Delta \mathcal{L} = \mathcal{L}_{\text{end}} - \mathcal{L}_{\text{start}}$. Meanwhile, we represent the average Loss of the last checkpoint as $\mathcal{L}_{\text{mean}}$.

Next, we can classify the tokens based on $\Delta \mathcal{L}$ and the $\mathcal{L}_{\text{mean}}$. We categorize tokens with $\Delta \mathcal{L} < -0.2$ as H→L (loss decreases from high to low) category tokens, and tokens with $\Delta \mathcal{L} > 0.2$ as L→H (loss increases from low to high) category tokens. If $-0.2 \leq \Delta \mathcal{L} \leq 0.2$ and $l_n \leq \mathcal{L}_{\text{mean}}$, then tokens are classified as L→L (loss remains low); if $l_n >\mathcal{L}_{\text{mean}}$, they are classified as H→H (loss remains high). 
In \autoref{fig:tokenppl_7b}, we have added the tokens' loss curves of the 7B model which is consistent with the other experimental settings in \autoref{sec:analysis:dynamics}, for readers to refer to whether similar phenomena exist on larger models. 
In \autoref{fig:hhll_case}, we visualize examples of the four categories of tokens in actual text.

\subsection{Non-Converging Tokens in Pretrainig}
\label{sec:appendix:non_converging_token}

In \autoref{sec:analysis:dynamics}, we mentioned that during the training process, only a minority of tokens belong to the H→L category. Among the remaining categories of H→H and L→L tokens, there are tokens that exhibit significant fluctuations during training. Furthermore, there are instances where H→L tokens are not effectively learned. Therefore, in our analysis, we specifically select those tokens from these categories that demonstrate considerable variability and distinct loss.
We visualize these tokens that exhibit abnormal behavior during the training process. As illustrated in \autoref{fig:token_fluctuating_example}, we find that the majority of these tokens originate from rather chaotic corpora. For instance, the corpora may include a mix of custom symbols, unintelligible gibberish, and information such as timetables and bibliographic references. Within a segment of normal text, there may also be fluctuations in the usage of common conjunctions, word suffixes, and punctuation marks. The latter may not necessarily be disastrous for training; in fact, it could represent a normal occurrence. However, if we can effectively mitigate the losses caused by the former, it might lead to more stable and efficient model training.

%% file: sec/appendix/a2_evalution.tex
\section{Evalution Details}
\label{sec:appendix:evalution}

\subsection{Math Evalution}
\label{sec:appendix:math_evalution}
We conducted a comprehensive evaluation of the model across various math reasoning benchmarks, encompassing a range of difficulties from elementary to university level, multiple mathematical domains, and diverse question types including multiple-choice and open-ended questions.
Our benchmarks include GSM8k~\citep{cobbe2021gsm8k}, MATH \citep{hendrycks2021measuring}, GSM-Hard \citep{gao2022pal}, SVAMP~\citep{patel2021svamp}, ASDIV \citep{miao-etal-2020-diverse},  MAWPS \citep{koncel-kedziorski-etal-2016-mawps}, TabMWP (TAB) \citep{lu2023dynamic}, MathQA (MQA) \citep{amini2019mathqa}, MMLU-STEM \citep{hendrycks2020measuring}, and SAT~\citep{azerbayev2023llemma}.

\subsection{General Evalution}
\label{sec:appendix:general_evalution}
In the evaluation of general domain, we followed the lm-evaluation-harness~\citep{eval-harness} and evalute model on MMLU~\citep{hendrycks2020measuring}, BBH~\citep{suzgun2022challenging}, AGIEval~\citep{zhong2023agieval}, ARC-Easy and ARC-Challenge~\citep{clark2018arc}, BoolQ~\citep{clark2019boolq}, PIQA~\citep{bisk2020piqa}, Hellaswag~\citep{zellers2019hellaswag}, WinoGrande~\citep{sakaguchi2021winogrande}, OpenBookQA~\citep{mihaylov2018obqa}. 
On HumanEval~\citep{zheng2023codegeex} and TydiQA~\citep{clark2020tydiqa}, we follow the evaluation pipeline of open-instrcut~\citep{ivison2023camels} and report {Pass@1} and {Pass@10} for HumanEval and {F1} for TydiQA. For MBPP~\citep{austin2021mbpp} benchmark, we follow the evaluation pipeline of DeepSeek-Coder~\citep{deepseek-coder}, and report {Pass@1} and {Pass@10}.

%% file: sec/appendix/a3_acc_loss.tex
\section{Relate the Selected Tokens' Loss to Downstream Task Performance}
\label{sec:appendix:acc_loss}

In this section, we declare the details about correlating the loss of selected tokens with the performance of downstream tasks. Concurrent study has explored similar methods to study the impact of scaling laws with the performance of models in downstream tasks~\citep{gadre2024LanguageMS}. Our analysis here differs in that it aims to elucidate the relationship between the decrease/increase in loss for selected/unselected tokens and the model's performance on downstream tasks. 

We use the average accuracy of MATH and GSM8K as the standard for measuring downstream tasks performance of model. Based on the trend of data points in \autoref{fig:acc_loss}, we propose the relationship between the average accuracy of downstream tasks and selected/unselected tokens' loss,

\begin{equation}
Acc(\mathcal{L}) = \log (a * \mathcal{L} + c)
\end{equation}

The parameters $a$ and $c$ are fitted from the data. If the loss of selected tokens $\mathcal{L}_s$ is used for fitting, then $a > 0$. Conversely, if the loss of unselected tokens $\mathcal{L}_{us}$ is used for fitting, then $a < 0$. Therefore, we believe that training the model on selected tokens can effectively improve its performance on downstream tasks, while unselected tokens may have a detrimental effect on the model's performance in downstream tasks.

%% file: sec/appendix/a4_examples_tokens_selected.tex
\section{Examples of Tokens Selected by \method{}}

\subsection{Token Selected Examples}
\label{sec:appendix:examples_token_selected}

In \autoref{fig:token_select_example}, we present several examples of tokens selected by the \method{} method, with content marked in \textcolor[RGB]{30,144,255}{blue} indicating the tokens actually chosen during the pretraining process.

\subsection{Dynamic Token Selected}
\label{sec:appendix:dynamic_token_selected}

In \autoref{fig:example_dynamic_token}, we display the dynamic changes in token selection tendencies throughout the \method{} training process. We chose four checkpoints during the training process (0\%, 33\%, 66\%, and 100\%) to analyze the current tendencies in token selection. The preferences for token selection are indicated by different colors, ranging from high to low preference, typically represented as \textcolor[RGB]{0,0,255}{deep blue}, \textcolor[RGB]{30,144,255}{blue}, black, \textcolor[RGB]{255,180,150}{orange}, and \textcolor[RGB]{255,100,0}{dark orange}, respectively.

%% file: sec/appendix/a5_self-reference_setting.tex
\section{Self-Reference Setting}
\label{sec:appendix:self-reference-setting}

\input{table/self_ref_results_appendix}

In this section, we will provide a detailed introduction to the reference loss score function and information entropy score function in SLM. Reference loss score function is to directly use the loss of the reference model as the basis for selecting tokens. The higher the token's loss of the reference model, the lower the expectation that the token will be selected. The score $\mathcal{L}_{\text{RM}}(x_i)$ can be directly obtained by referring to \autoref{equ:ref_loss}. Information entropy score function is to select the corresponding token based on the information entropy of the reference model in each token. The information entropy of token $x_i$ can be expressed as:

\begin{equation}
\mathcal{H}_{\text{RM}}(x_i) = -\sum_{k=1}^{V} P(t_k|x_{<i}) \log P(t_k|x_{<i}),
\end{equation}

where $t_k$ represents the i-th token in the vocabulary, and $V$ represents the size of the vocabulary. The intuition of this strategy is that the higher the information entropy, the higher the uncertainty of the token in the context. Therefore, we consider that if the language model is still uncertain for certain tokens after pretraining, we do not expect that the language model will learn it during pretraining. In \autoref{tab:self-ref-1b-slm}, we provide more SLM results, including different select ratios and combinations of two score functions, for the convenience of the readers to refer to.

%% file: table/self_ref_results_appendix.tex
\begin{table*}[t]
\centering
\small
\setlength{\tabcolsep}{2pt}
\caption{\textbf{Full Self-Reference results on Tinyllama-1.1B.}}
\label{tab:self-ref-1b-slm}
\resizebox{\linewidth}{!}{
\begin{tabular}{c|c|cccccc|c}
\toprule
\multicolumn{1}{c|}{\textbf{\begin{tabular}[c]{@{}c@{}}Score\\ Function\end{tabular}}} & \multicolumn{1}{c|}{\textbf{\begin{tabular}[c]{@{}c@{}}Select\\ Ratio\end{tabular}}} & \textbf{GSM8K} & \textbf{MATH} & \textbf{SVAMP} & \textbf{ASDiv} & \textbf{MAWPS} & \textbf{MQA} & \textbf{AVG} \\
\midrule
- & 100\% & 6.3 & 2.6 & 21.7 & 36.7 & 47.7 & 13.9 & 21.5 \\
\midrule
\multirow{5}{*}{$\mathcal{L}_{\text{RM}}(x_i)$} & 90\% & \textbf{7.4} & 4.4 & \textbf{23.4} & 38.7 & 51.9 & \textbf{14.4} & 23.4 \\
& 80\% & 6.4 & \textbf{4.6} & 23.1 & 39.7 & 52.0 & 14.3 & 23.4 \\
& 70\% & 6.7 & \textbf{4.6} & 23.3 & \textbf{40.0} & \textbf{54.5} & 14.3 & \textbf{23.9} \\
& 60\% & 7.0 & \textbf{4.6} & 22.2 & 38.5 & 52.2 & 13.7 & 23.0 \\
& 50\% & 5.7 & 4.2 & 20.7 & 36.7 & 46.7 & 10.3 & 20.7  \\
\midrule

\multirow{5}{*}{$\mathcal{H}_{\text{RM}}(x_i)$}
& 90\% & 6.7 & 3.0 & 23.7 & 40.3 & 52.3 & 13.1 & 23.2 \\
& 80\% & 6.8 & 3.6 & 22.5 & \textbf{40.6} & \textbf{52.9} & 13.6 & 23.3 \\
& 70\% & \textbf{7.0} & 4.8 & 23.0 & 39.3 & 50.5 & 13.5 & 23.0 \\
& 60\% & 6.5 & 4.8 & \textbf{26.5} & 37.3 & 49.7 & \textbf{15.6} & \textbf{23.4} \\
& 50\% & 4.7 & \textbf{5.8} & 20.9 & 33.8 & 42.5 & 11.1 & 19.8  \\
\midrule

\multirow{5}{*}{$\mathcal{H}_{\text{RM}}(x_i)\cup\mathcal{L}_{\text{RM}}(x_i)$}
& $50\% \cup 70\% (80\%)$& 6.4 & 3.6 & 22.7 & 38.4 & 52.6 & 15.3 & 23.2 \\
& $70\% \cup 60\% (77\%)$& 6.3 & 4.6 & 24.4 & 39.6 & 51.4 & \textbf{16.3} & \textbf{23.8} \\
& $70\% \cup 50\% (75\%)$& 6.9 & 5.6 & 23.2 & \textbf{39.9} & \textbf{52.9} & 12.6 & 23.5 \\
& $60\% \cup 60\% (70\%)$& 6.7 & 5.2 & \textbf{24.7} & 39.2 & 50.6 & 14.6 & 23.5 \\
& $60\% \cup 50\% (68\%)$& 7.1 & 5.8 & 21.7 & 37.3 & 49.6 & 15.3 & 22.8 \\
& $60\% \cup 40\% (65\%)$& \textbf{7.3} & \textbf{6.0} & 23.6 & 36.9 & 48.6 & 13.1 & 22.6 \\
\midrule

\multirow{5}{*}{$\mathcal{H}_{\text{RM}}(x_i)\cap\mathcal{L}_{\text{RM}}(x_i)$} 
& $80\% \cap 90\% (76\%)$& 6.0 & 4.4 & 23.7 & 38.5 & 51.2 & 13.3 & 22.8 \\
& $75\% \cap 75\% (72\%)$& 7.8 & 5.2 & \textbf{24.2} & 39.4 & \textbf{54.9} & 14.7 & 24.4 \\
& $70\% \cap 90\% (68\%)$& 6.8 & 4.6 & 22.2 & 40.3 & 53.0 & 14.8 & 23.6 \\
& $80\% \cap 80\% (67\%)$& \textbf{8.2} & \textbf{6.4} & 21.2 & 39.1 & 53.4 & 15.0 & 23.9 \\
& $70\% \cap 70\% (60\%)$& 7.1 & 5.0 & 23.5 & \textbf{41.2} & 53.8 & \textbf{18.0} & \textbf{24.8} \\
\bottomrule
\end{tabular}
}
\end{table*}

%% file: sec/appendix/a6_weak-to-strong.tex
\section{Weak-to-Strong Generalization}

\input{table/small_ref_results}
Apart from the main experiments where we use the same base model for the reference and continual pretraining, we also investigate if a smaller reference model can effectively guide the pretraining of a larger model.
We use Tinyllama-1.1B as reference model and continual pretraining Llama-2-7B on 15B OpenWebMath tokens.
Results presented in \autoref{tab:small-ref-result} indicate that, despite the considerable gap between the small and large models~\citep{contrastivedecoding23}, employing the small reference model to token selection can still yield benefits to the pre-training of the larger model.
If reference and training models have different vocabularies, one can consider performing token alignment \citep{wan2024knowledge, fu2023specializing}, which we leave for future work.

%% file: table/small_ref_results.tex
\begin{table*}[t]
\centering
\setlength{\tabcolsep}{2pt}
\caption{\textbf{Weak-to-Strong generalization result on math benchmark.}
}
\label{tab:small-ref-result}
\resizebox{\linewidth}{!}{
\begin{tabular}{lc|ccccccccc|c}
\toprule
\textbf{Model} & \textbf{Train Toks} & \textbf{GSM8K} & \textbf{MATH} & \textbf{SVAMP} & \textbf{ASDiv} & \textbf{MAWPS} & \textbf{TAB}& \textbf{MQA} & \multicolumn{1}{c}{\textbf{\begin{tabular}[c]{@{}c@{}}MMLU\\ STEM\end{tabular}}} & \textbf{SAT} & \textbf{AVG} \\
\midrule
Llama-2-7B-CT & 15B & 28.4 & 13.6 & 50.3 & 62.8 & 79.5 & 37.6 & 34.1 & 41.6 & 43.5 & 43.5 \\
Llama-2-7B-CT w/ 1B RM & 10.5B &  29.8 & 16.0 & 55.5 & 63.7 & 80.4 & 37.9 & 34.3 & 38.2 & 43.8 & 44.4 \\
\bottomrule
\end{tabular}
}
\end{table*}

%% file: sec/appendix/a7_example_case.tex
\begin{figure}[t]
\centering
\includegraphics[width=\textwidth]{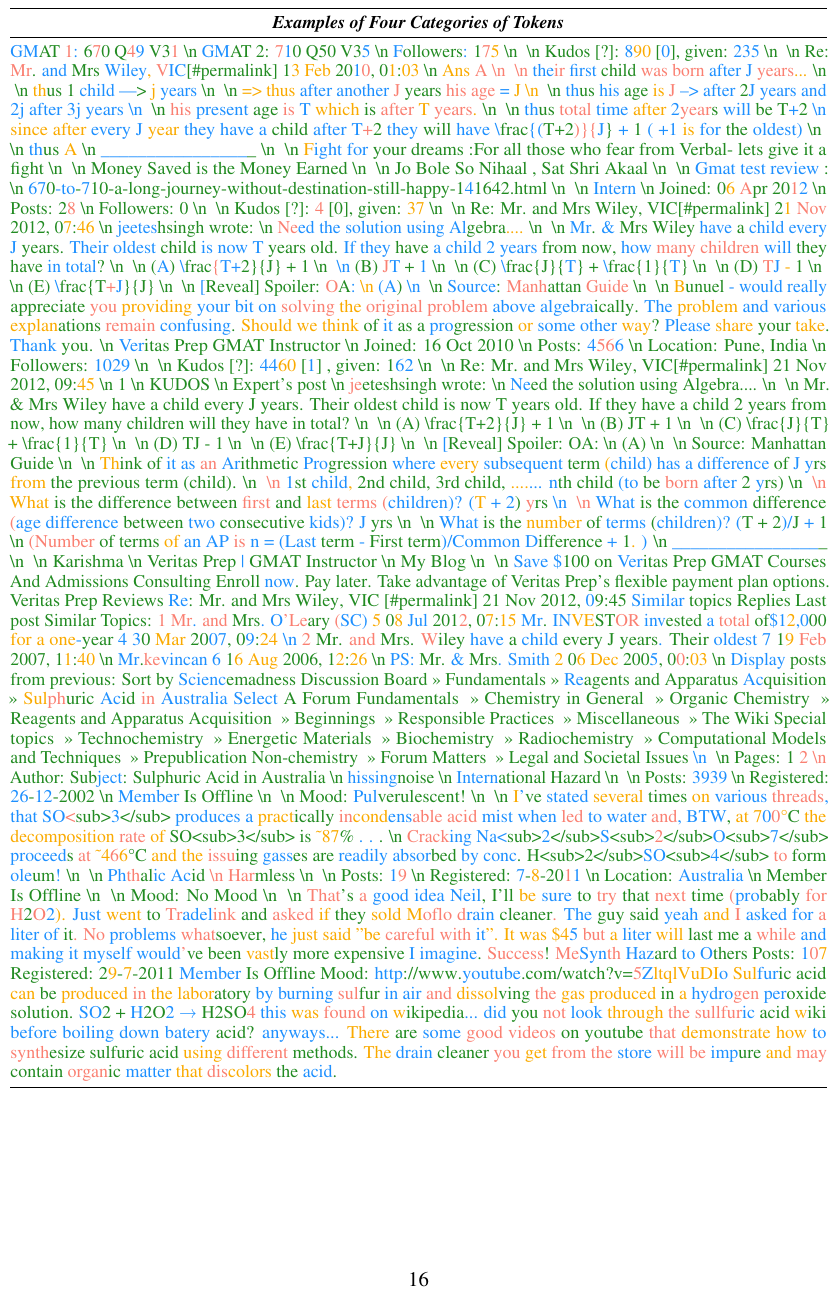}
\caption{\textbf{Sample text containing four categories of tokens.} Among them, \textcolor[RGB]{30,144,255}{blue} represents tokens of categorie H→L, \textcolor[RGB]{34,139,34}{green} indicates tokens of categorie L→L, \textcolor[RGB]{250,170,0}{yellow} signifies tokens of categorie H→H, and \textcolor[RGB]{250,128,114}{red} denotes tokens of categorie L→H.}
\label{fig:hhll_case}
\end{figure}

\begin{figure}[t]
\centering
\includegraphics[width=\textwidth]{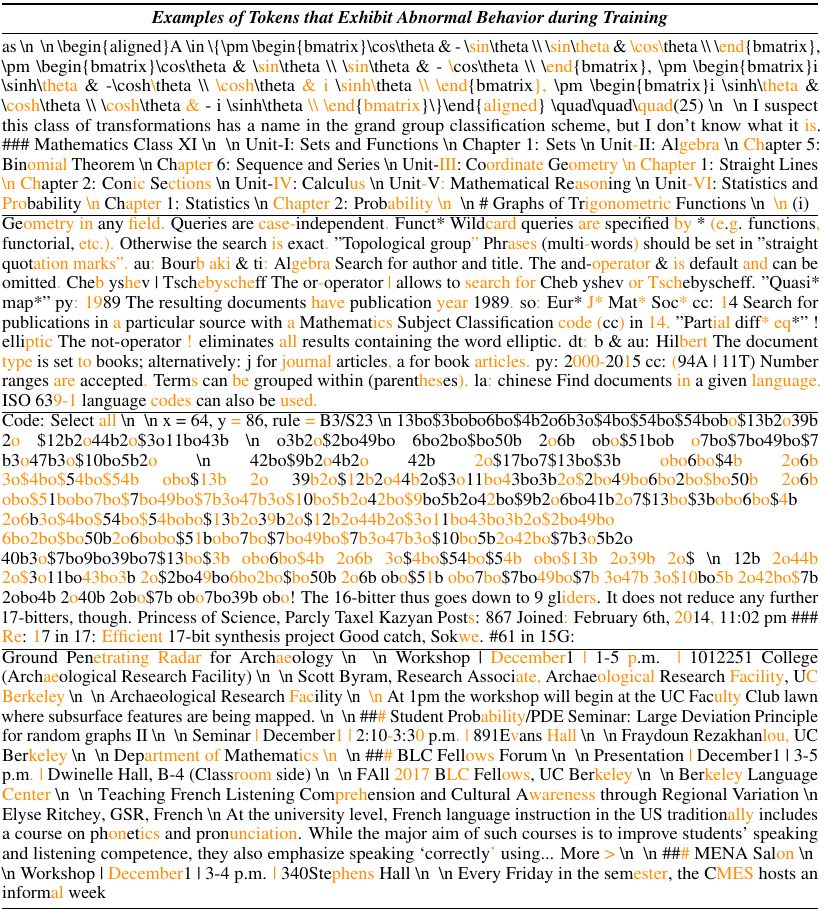}
\caption{\textbf{An example of an abnormal state of token perplexity during pretrainig process.} The tokens highlighted in \textcolor[RGB]{255,140,0}{orange} represent tokens that were significant abnormalities during the pretraining process.}
\label{fig:token_fluctuating_example}
\end{figure}

\begin{figure}[t]
\centering
\includegraphics[width=\textwidth]{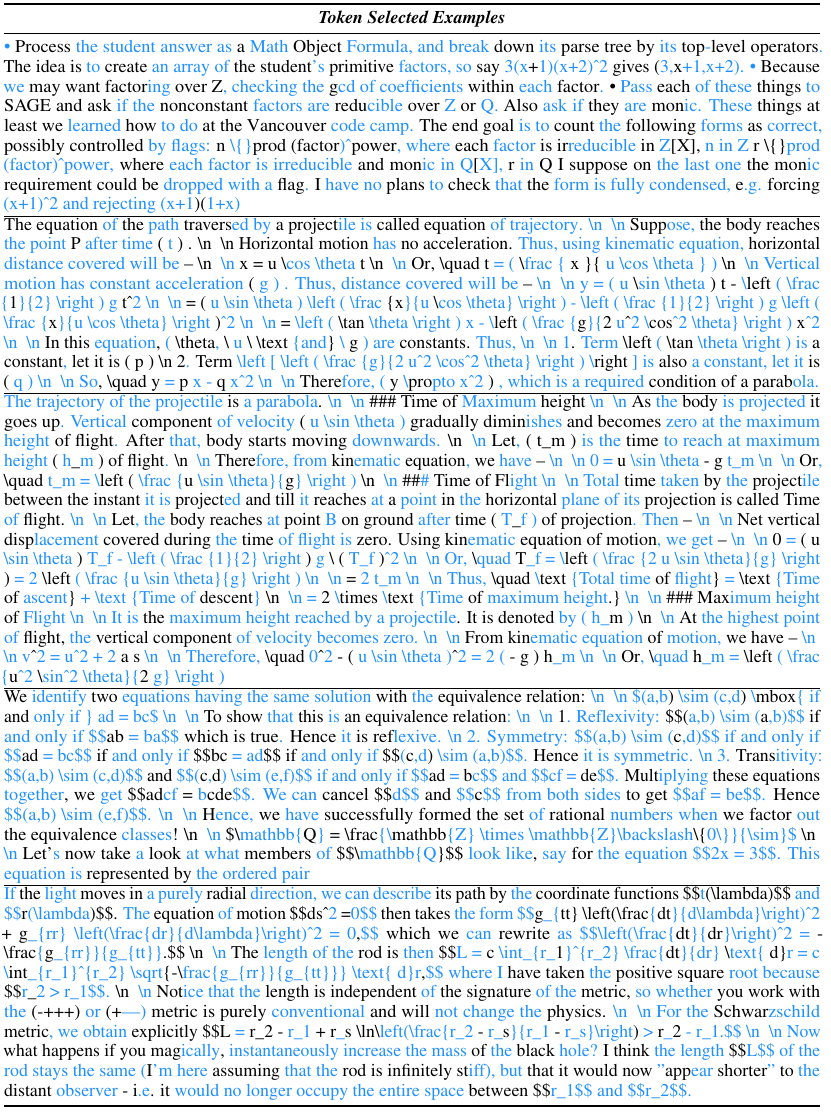}
\caption{\textbf{Specific examples of selecting tokens during the selective pretraining process of the \model{}.} The tokens marked in \textcolor[RGB]{30,144,255}{blue} represent the actual tokens trained during the training process, while the remaining black tokens are not trained during the training process.}
\label{fig:token_select_example}
\end{figure}

\begin{figure}[t]
\centering
\includegraphics[width=\textwidth]{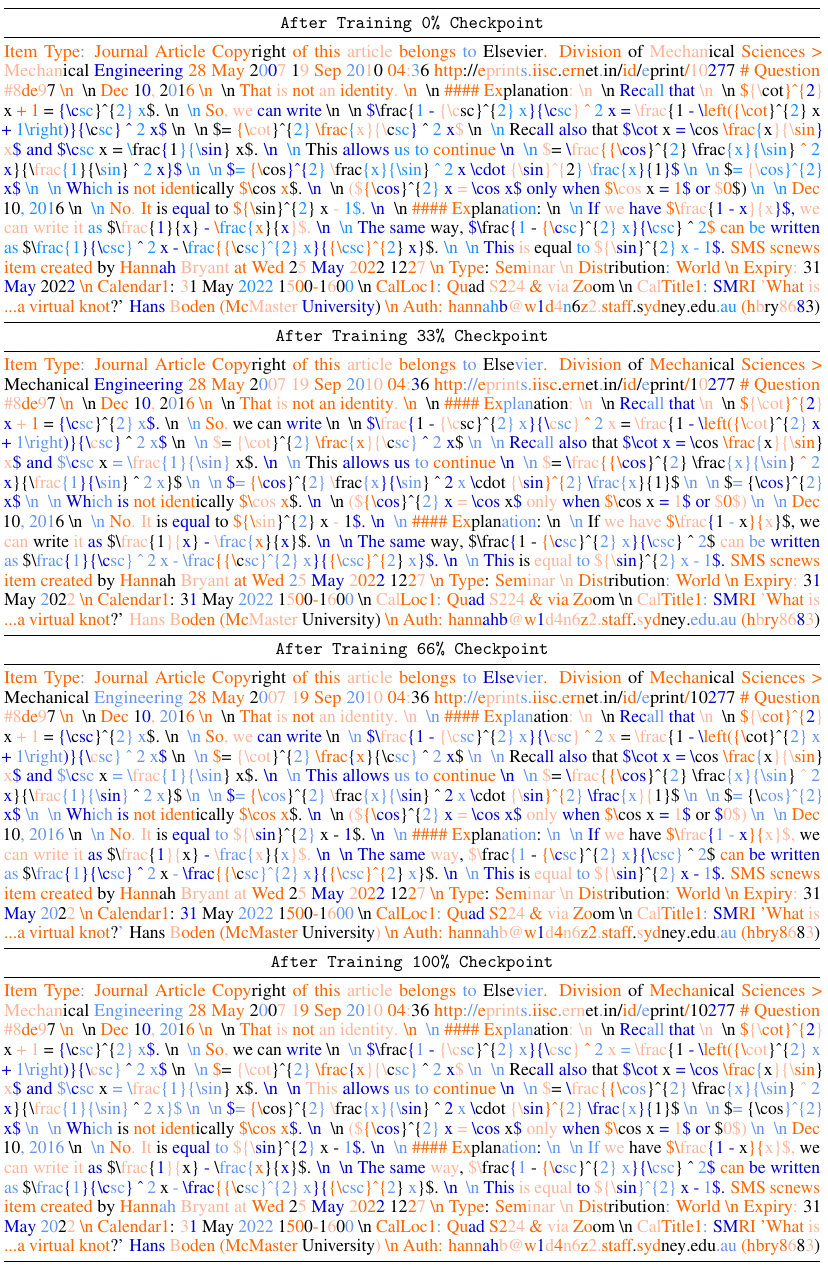}
\caption{\textbf{An example of dynamic token selection changes during the training process}, which illustrated with five different score levels represented by \textcolor[RGB]{0,0,255}{deep blue}, \textcolor[RGB]{30,144,255}{light blue}, black, \textcolor[RGB]{255,180,150}{light orange}, and \textcolor[RGB]{255,100,0}{dark orange}. The bluer the color indicates a higher tendency for the token to be selected, while the more orange the color suggests a lower tendency for the token to be selected.}
\label{fig:example_dynamic_token}
\end{figure}